\documentstyle[11pt,
	       lingmacros,
               mylingmacros,
               fullname]{article}
\input lfgmacros
\input tree-dvips

\makeatletter
\def \lexentry #1#2#3{\setbox1\hbox{#1}\setbox2\hbox{#2}%
\ifdim \wd1 >3em \lexentrydimen=\wd1 %
\else \lexentrydimen=3em \fi%
\ifdim \wd2 >1em \lexcategorydimen=\wd2 %
\else \lexcategorydimen=1em \fi%
\noindent%
\makebox[\lexentrydimen][l]{\box1}\hspace*{1em}\nobreak%
\makebox[\lexcategorydimen][l]{\box2}\hspace*{1em}\nobreak%
\feqs{#3}}

\def\section{\@startsection {section}{1}{\z@}{-3.5ex plus -1ex minus
 -.2ex}{2.3ex plus .2ex}{\normalsize\bf}}
\def\subsection{\@startsection{subsection}{2}{\z@}{-3.25ex plus -1ex minus
 -.2ex}{1.5ex plus .2ex}{\normalsize\bf}}
\def\subsubsection{\@startsection{subsubsection}{3}{\z@}{-3.25ex plus
-1ex minus -.2ex}{1.5ex plus .2ex}{\normalsize\bf}}
\def\paragraph{\@startsection
 {paragraph}{4}{\z@}{3.25ex plus 1ex minus .2ex}{-1em}{\normalsize\bf}}

\def\eqalign#1{\null\,\vcenter{\openup\jot\m@th
  \ialign{\strut\hfil$\displaystyle{##}$&$\displaystyle{{}##}$\hfil
      \crcr#1\crcr}}\,}

\newcommand{\lam}[1]{\lambda #1. }

\newcommand{\linimp}{\;\mbox{$-\hspace*{-.4ex}\circ$}\;}
\newcommand{\means}{\makebox[1.2em]{$\leadsto$}}
\newcommand{\meansub}[1]{\makebox[1.5em]{$\leadsto_{#1}$}}
\newcommand{\IT}[1]{\mbox{\it #1\/}}
\newcommand{\BF}[1]{\mbox{\bf #1}}
\newcommand{\intn}{\hat{\ }}
\newcommand{\extn}{\check{\ }}
\newcommand{\oneover}[2]{\begin{array}{@{}c@{}} #1 \\
\hline #2  \end{array}}
\newcommand{\pex}[1]{(\ref{#1})}
\newcommand{\attr}[2]{\mbox{$(#1\;#2)$}}
\newcommand{\Up}{\mbox{$\uparrow$}}
\newcommand{\Down}{\mbox{$\downarrow$}}
\newcommand{\Ups}{\Up_\sigma}
\newcommand{\All}[1]{\forall #1.\;}
\newcommand{\Al}{\IT{Al}\,}

\def\fd#1{\setlength{\baselineskip}{0pt}
  \small
     \hbox{#1}}
\def\fdx#1{\setlength{\baselineskip}{0pt}\vcenter{#1}}
\def\fdand#1{\(\left[\fdx{#1}\right]\)}

\def\feat#1#2{\vskip .4ex\hbox{\hspace{.2em}#1\hspace{1em}#2\hspace{.2em}}%
\vskip .8ex}

\def\var{\mbox{\sc var}}
\def\restr{\mbox{\sc restr}}
\def\ant{\mbox{\sc ant}}
\def\alt{{\tt\char`\|}}
\makeatother

\title{Quantifiers, Anaphora, and Intensionality}
\author{Mary Dalrymple\thanks{Xerox PARC, Palo Alto CA 94304;
{\tt \{dalrymple,lamping,saraswat\}@parc.xerox.com}}\\
John Lamping\footnotemark[1]\\
Fernando Pereira\thanks{AT\&T Bell Laboratories, Murray Hill NJ 07974;
{\tt pereira@research.att.com}}\\
Vijay Saraswat\footnotemark[1]}

\begin{document}
\maketitle

The relationship between Lexical-Functional Grammar (LFG) {\em
functional structures} (f-structures) for sentences and their semantic
interpretations can be expressed directly in a fragment of linear
logic in a way that correctly explains the constrained interactions
between quantifier scope ambiguity, bound anaphora and intensionality.

The use of a deductive framework to account for the compositional
properties of quantifying expressions in natural language obviates the
need for additional mechanisms, such as Cooper storage, to represent
the different scopes that a quantifier might take.  Instead, the
semantic contribution of a quantifier is recorded as a
logical formula whose use in a proof will establish the scope of the
quantifier. Different proofs will in general lead to different scopes.
In each complete proof, the properties of linear logic will ensure
that each quantifier is properly scoped.

The interactions between quantified NPs and intensional verbs such as
`seek' are also accounted for in this deductive setting. A single
specification in linear logic of the argument requirements of
intensional verbs is sufficient to derive the correct reading
predictions for intensional-verb clauses both with nonquantified
and with quantified direct objects. In particular, both {\em de
dicto}\/ and {\em de re}\/ readings are derived for quantified
objects. The effects of type-raising or quantifying-in rules in other
frameworks here just follow as linear-logic theorems.

While our approach resembles current categorial approaches in
important ways
\cite{Moortgat:categorial,Moortgat:discontinuous,Morrill:type-logical,%
Carpenter:quant-scope},
it differs from them in allowing the greater type flexibility of
categorial semantics \cite{VanBenthem:LgInAction} while maintaining a
precise connection to syntax. As a result, we are able to provide
derivations for certain readings of sentences with intensional verbs
and complex direct objects that are not derivable in current purely
categorial accounts of the syntax-semantics interface.

\section{Introduction}

This paper describes a part of our ongoing investigation in the use of
formal deduction to explicate the relationship between syntactic
analyses in Lexical-Functional Grammar (LFG) and semantic
interpretations.  We use {\em linear logic\/}
\cite{Girard:Linear} to represent the connection between two
dissimilar linguistic levels: LFG f-structures and their semantic
interpretations.

F-structures provide a uniform representation of syntactic information
relevant to semantic interpretation that abstracts away from the
varying details of phrase structure and linear order in particular
languages.  As \namecite{Halvorsen:Sitsem} notes, however, the
flatter, unordered functional structure of LFG does not fit well with
traditional semantic compositionality, based on functional abstraction
and application, which mandates a rigid order of semantic composition.
We are thus led to a more relaxed form of compositionality, in which,
as in more traditional ones, the semantics of each lexical entry in a
sentence is used exactly once in interpretation, but without imposing
a rigid order of composition. Approaches to semantic interpretation
that encode semantic representations in attribute-value structures
\cite{PollardSag:HPSG1,FenstadEtAl:SitLgLogic,PollardSag:HPSG2} offer
such a relaxation of compositionality, but are unable to properly
represent constraints on variable binding and scope
\cite{Pereira:SemComp}.

The present approach, in which linear logic is used to specify the
relation between f-structures and their meanings, provides exactly
what is required for a calculus of semantic composition for LFG. It
can directly represent the constraints on the creation and use of
semantic units in sentence interpretation, including those pertaining
to variable binding and scope, without forcing a particular
hierarchical order of composition, except as required by the
properties of particular lexical entries.

The use of formal deduction in semantic interpretation was implicit in
deductive systems for categorial syntax
\cite{Lambek:SentStruct}, and has been made explicit through
applications of the Curry-Howard parallelism between proofs and terms
in more recent work on categorial semantics
\cite{vanBenthem:lambek,VanBenthem:LgInAction}, labeled deductive
systems \cite{Moortgat:Labelled} and flexible categorial systems
\cite{Hendriks:flexibility}. Accounts of the syntax-semantics
interface in the categorial tradition require that syntactic and
semantic analyses be formalized in parallel algebraic structures of
similar signatures, based on generalized application and abstraction
(or residuation) operators and structure-preserving relations between
them.  Those accounts therefore force the adoption of categorial
syntactic analyses, with an undesirably strong dependence on phrase
structure and linear order.

We have previously shown that the linear-logic formalization of the
syntax-semantics interface for LFG provides simple and general
analyses of modification, functional completeness and coherence, and
complex predicate formation~\cite{DLS:EACL,DHLS:ROCLING}.  In the
present paper, the analysis is extended to the interpretation of
quantified NPs.  After an overview of the approach, we present our
analysis of the compositional properties of quantified NPs, and we
show that the analysis correctly accounts for scope ambiguity and its
interactions with bound anaphora.  We also present an analysis of
intensional verbs, which take quantified arguments, and show that our
approach predicts the full range of acceptable readings without
appealing to additional machinery.

\section{LFG and Linear Logic}

\paragraph{Syntactic framework}

LFG assumes two syntactic levels of representation.  Constituent
structure ({\em c-structure}) encodes phrasal dominance and precedence
relations, and is represented as a phrase structure tree.  Functional
structure ({\em f-structure}) encodes syntactic predicate-argument
structure, and is represented as an attribute-value matrix.  The
c-structure and f-structure for sentence \pex{ex:bah} are given in
\pex{ex:bahfs}:

\enumsentence{\label{ex:bah} Bill appointed Hillary.}

\enumsentence{\label{ex:bahfs}\evnup{
\begin{tabular}{cc}
C-structure: & F-structure: \\[1em]
\modsmalltree{3}{\mc{3}{\node{l}{S}\hspace*{1em}}\\
            \node{m}{NP}&\mc{2}{\hspace*{1em}\node{n}{VP}}\\
                        &\node{p}{V}&\node{t}{NP}\\
            \node{q}{Bill}&\node{r}{appointed}&\node{s}{Hillary}}
\nodeconnect{l}{m}
\nodeconnect{l}{n}
\nodeconnect{m}{q}
\nodeconnect{n}{p}
\nodeconnect{n}{t}
\nodeconnect{t}{s}
\nodeconnect{p}{r}
&
\fd{
\fdand{\feat{\pred}{`{\sc appoint}'}
       \feat{\subj}{\fdand{\feat{\pred}{`{\sc Bill}'}}}
       \feat{\obj}{\fdand{\feat{\pred}{`{\sc Hillary}'}}}}}
\end{tabular}}}

\noindent As illustrated, an f-structure consists of a collection
of attributes, such as \pred, \subj, and \obj, whose values can, in
turn, be other f-structures.

The relationship between c-structure trees and the corresponding
f-structures is given by a {\em functional projection\/} function $\phi$
from c-structure nodes to f-structures. More generally, LFG analyses
involve several levels of linguistic representation called {\em
projections} related by means of {\em projection functions\/}
\cite{Kaplan:3Sed,HalvorsenKaplan:Projections}.  For instance,
phonological, morphological, or discourse structure might be
represented by a phonological, morphological, or discourse projection,
related to other projections by means of functional specifications.

The functional projection of a c-structure node is the solution of
constraints associated with the phrase-structure rules and lexical
entries used to derive the node. In each rule or lexical entry
constraint, the \Up\ metavariable refers to the $\phi$-image of the
mother c-structure node, and the \Down\ metavariable refers to the
$\phi$-image of the nonterminal labeled by the constraint
\cite[page~183]{KaplanBresnan:LFG}. For example, the following
annotated phrase-structure rules were used in the analysis of sentence
(\ref{ex:bah}):

\enumsentence{
\phraserule{S}{
\rulenode{NP\\ \attr{\Up}{\subj} = \Down}
\rulenode{VP\\ \up = \Down}}}

\noindent The annotations on the rule indicate that the f-structure for
the S (\Up\ in the annotation on the NP node) has a \subj\ attribute
whose value is the f-structure for the NP daughter (\Down\ in the
annotation on the NP node), and that the S node corresponds to an
f-structure which is the same as the f-structure for the VP daughter.

When the phrase-structure rule for S is used in the analysis of a
particular sentence, the metavariables \Up\ and \Down\ are
instantiated to particular f-structures placed in correspondence with
nodes of the c-structure.  We will refer to actual f-structures by
giving them names such as $f$, $g$, and $h$.  The instantiated phrase
structure rule is given in (\ex{1}), with the $\phi$ correspondence
between c-structure nodes and f-structures indicated by the directed
arcs from phrase-structure nodes to attribute-value matrices:

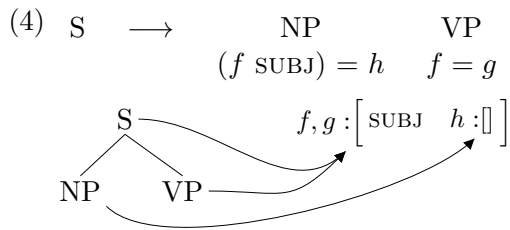
\vbox{\enumsentence{\evnup{
\phraserule{S}{
\rulenode{NP\\ ($f$ \subj) = $h$}
\rulenode{VP\\ $f = g$}}}

\modsmalltree{3}{\mc{3}{\node{l}{S}\hspace*{1em}}\\
            \node{m}{NP}&\mc{2}{\hspace*{1em}\node{n}{VP}}}
\hspace*{2em}%
\fd{$f,g:$\node{a}{\fdand{
       \feat{\SUBJ}{$h:$\node{b}{\fdand{ }}}}}}
\anodecurve[r]{l}[bl]{a}{3em}
\anodecurve[r]{n}[bl]{a}{3em}
\anodecurve[br]{m}[bl]{b}{3em}
\nodeconnect{l}{m}
\nodeconnect{l}{n}}}

Lexical entries also use the metavariables \Up\ and \Down\ to encode
information about the f-structures of the preterminal nodes that
immediately dominate them.  A partial lexical entry for the word
`Bill' is:

\enumsentence{
\lexentry{Bill}{NP}{\attr{\Up}{\pred} = `{\sc Bill}'}}

\noindent This entry states that `Bill' has syntactic category NP.
The constraint \mbox{\attr{\Up}{\pred} = `{\sc Bill}'} states that
the preterminal node immediately dominating the terminal symbol `Bill'
has an f-structure whose value for the attribute \pred\ is `{\sc
Bill}'.  In this paper, we will provide only the most minimal
f-structural representations, leaving aside all details of syntactic
specification; in this example, for instance, agreement and other
syntactic features of `Bill' have been omitted.

For a particular instance of use of the word `Bill', the following
c-structure and f-structure configuration results:

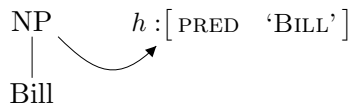
\vbox{\enumsentence{($h$ \pred) = `{\sc Bill}'\\[2em]
\modsmalltree{1}{\node{l}{NP}\\
            \node{m}{Bill}}
\hspace*{2em}%
\fd{$h:$\node{a}{\fdand{
       \feat{\pred}{`{\sc Bill}'}}}}
\anodecurve[br]{l}[bl]{a}{2em}
\nodeconnect{l}{m}}}

\noindent Other lexical entries similarly specify features of
the f-structure of the immediately dominating preterminal node.  The
following is a list of the phrase structure rules and lexical entries
used in the analysis of example (\ref{ex:bah}):\footnote{Those
familiar with other analyses within the LFG framework will notice that
we have not included a list of grammatical functions subcategorized
for by the verb `appoint'; this is because we assume a different
treatment of the LFG requirements of completeness and coherence.  We
return to this point below.}

\vbox{\enumsentence{
\phraserule{S}{
\rulenode{NP\\ \attr{\Up}{\subj} = \Down}
\rulenode{VP\\ \up = \Down}}\\[1em]
\phraserule{VP}{
\rulenode{V\\ \up = \Down}
\rulenode{NP\\ \attr{\Up}{\obj} = \Down}}}}

\vbox{\enumsentence{\label{ex:synlex}
\lexentry{Bill}{NP}{
\attr{\Up}{\pred} = `{\sc Bill}'}

\lexentry{appointed}{V}{
\attr{\Up}{\pred}= `{\sc appoint}'}

\lexentry{Hillary}{NP}{
\attr{\Up}{\pred} = `{\sc Hillary}'}}}

\noindent For a more complete explication of the syntactic assumptions
of LFG, see \namecite{Bresnan:MentalRep}, \namecite{LRZ:LFG}, and the
references cited there.

\paragraph{Lexically-specified semantics}

A distinguishing feature of our work (and of other work within the LFG
framework) is that semantic composition does not take syntactic
dominance and precedence relations as the main input.  Instead, we
follow other work in LFG (Kaplan and Bresnan 1982, Halvorsen 1983,
Halvorsen and Kaplan 1988, and many others) in assuming that the
functional syntactic information encoded by f-structures determines
semantic composition.  That is, we believe that meaning composition is
mainly determined by syntactic relations such as {\em subject-of}\/,
{\em object-of}\/, {\em modifier-of}\/, and so on. Those relations are
realized by different c-structure forms in different languages, but
are represented directly and uniformly in the f-structure.

In LFG, syntactic predicate-argument structure is projected from
lexical entries.  Therefore, its effect on semantic composition will
for the most part -- in fact, in all the cases considered in this
paper -- be determined by lexical entries, not by phrase-structure
rules.  In particular, the two phrase-structure rules given above for
S and VP need not encode semantic information, but only specify how
grammatical functions such as \subj\ are expressed in English.  In
some cases, the constituent structure of a syntactic construction may
make a direct semantic contribution, as when properties of the
construction as a whole and not just of its lexical elements are
responsible for the interpretation of the construction. Such cases
include, for instance, relative clauses with no complementizer (`the
man Bill met').  We will not discuss construction-specific
interpretation rules in this paper.

In the same way as the functional projection function $\phi$
associates f-structures to c-structures as described above, we will
use a {\em semantic\/} or $\sigma$-{\em projection\/} function $\sigma$ to
map f-structures to {\em semantic\/} or $\sigma$-{\em structures\/}
encoding information about f-structure meaning. For instance, the
following lexical entry for `Bill' extends (\ref{ex:synlex}) with a
suitable constraint on semantic structure:
\enumsentence{\label{ex:lex-bill} \lexentry{Bill}{NP}{
\attr{\Up}{\pred} = `{\sc Bill}'\\ $\Ups \means \IT{Bill}$}}
\noindent The additional constraint $$\Ups \means \IT{Bill}$$ is what we
call the {\em meaning constructor\/} of the entry.  The expression
$\Ups$ stands for the {\em $\sigma$ projection}\/ of the f-structure
\Up.  The $\sigma$ projection is an attribute-value matrix like the
f-structure.  For simple entries such as this, the $\sigma$ projection
has no internal structure; below, we will examine cases in which the
$\sigma$ projection is structured with several different attributes.

As above, for a particular use of `Bill', the metavariable
\Up\ will be replaced by a particular f-structure $h$, with semantic
projection $h_\sigma$:

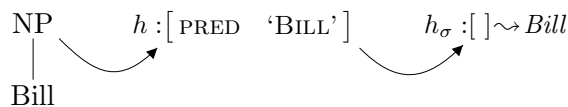
\vbox{\enumsentence{($h$ \pred) = `{\sc Bill}'\\[2em]
\modsmalltree{1}{\node{l}{NP}\\
            \node{m}{Bill}}
\hspace*{2em}%
\fd{$h:$\node{a}{\fdand{
       \feat{\pred}{`{\sc Bill}'}}}}
\hspace*{2em}%
\fd{$h_\sigma:$\node{b}{[\ ]}}$\means \IT{Bill}$
\anodecurve[br]{l}[bl]{a}{2em}
\anodecurve[br]{a}[bl]{b}{2em}
\nodeconnect{l}{m}}}

\noindent More generally, the association between the semantic
structure $h_\sigma$ and a meaning $P$ is represented by the atomic
formula $h_\sigma \means P$, where $\means$ is an otherwise
uninterpreted binary predicate symbol.  (In fact, we use not one but a
family of relations~$\meansub{\tau}$ indexed by the semantic type of
the intended second argument, although for simplicity we will omit the
type subscript whenever it is determinable from context.)  We can now
explain the meaning constructor in
\pex{ex:lex-bill}.  If a particular occurrence of `Bill' in a sentence
is associated with f-structure $h$, the syntactic constraint in the
lexical entry {\em Bill\/} will be instantiated as: $$\attr{h}{\pred} =
\mbox{`{\sc Bill}'}$$ and the semantic constraint will be instantiated as:
$$h_\sigma \means \IT{Bill}$$ representing the association between
$h_\sigma$ and the constant $\IT{Bill}$ representing its meaning.

We will often informally say that $P$ is $h$'s meaning without
referring to the role of the semantic structure $h_\sigma$ in
$h_\sigma \means P$. We will see, however, that f-structures and their
semantic projections must be distinguished, because semantic
projections can carry more information than just the association to
the meaning for the corresponding f-structure.

\paragraph{Logical representation of semantic compositionality}
We now turn to an examination of the lexical entry for `appointed'.
In this case, the meaning constructor is more complex, as it relates
the meanings of the subject and object of a clause to the clause's
meaning:

\enumsentence{\label{ex:appointed-lex}
\lexentry{appointed}{V}{
\attr{\Up}{\pred}= `{\sc appoint}'\\
$\All{ X, Y}\attr{\Up}{\subj}_\sigma\means X \otimes
\attr{\Up}{\obj}_\sigma\means Y \linimp\Ups \means \IT{appoint}\/(X, Y)$}}

\noindent The meaning constructor is the linear-logic formula:
$$\All{ X, Y}\attr{\Up}{\subj}_\sigma\means X \otimes
\attr{\Up}{\obj}_\sigma\means Y \linimp \Ups \means \IT{appoint}\/(X,
Y)$$ in which the linear-logic connectives of multiplicative
conjunction $\otimes$ and linear implication $\linimp$ are used to
specify how the meaning of a clause headed by the verb is composed
from the meanings of the arguments of the verb. For the moment, we can
think of the linear connectives as playing the same role as the
analogous classical connectives conjunction $\wedge$ and implication
$\rightarrow$, but we will soon see that the specific properties of the
linear connectives are essential to guarantee that lexical entries
bring into the interpretation process all and only the information
provided by the corresponding words.

The meaning constructor for `appointed' asserts, then, that if the
subject (\subj) of a clause with main verb `appointed' means $X$ and
its object (\obj) means $Y$, then the whole clause means
$\IT{appoint}(X,Y)$.\footnote{In fact, we believe that the correct treatment of
the relation between a verb and its arguments requires the use of {\em
mapping principles}\/ specifying the relation between the array of
semantic arguments required by a verb and their possible syntactic
realizations \cite{BresnanKanerva:Locative,Alsina:PhD,Butt:PhD}.  A
verb like `appoint', for example, might specify that one of its
arguments is an agent and the other is a theme.  Mapping principles
would then specify that agents can be realized as subjects and themes
as objects.

Here we make the simplifying assumption (valid for English)
that the arguments of verbs have already been
linked to syntactic functions and that this linking is represented in
the lexicon.  In the case of {\em complex predicates}\/ this
assumption produces incorrect results, as shown by
\namecite{Butt:PhD} for Urdu.  Mapping principles are very naturally
incorporated into the framework discussed here; see
\namecite{DLS:EACL} and \namecite{DHLS:ROCLING} for discussion and
illustration.}  The meaning constructor can thus be thought of as a
linear definite clause, with the variables $X$ and $Y$ playing the
same role as Prolog variables.

A particular instance of use of `appointed' produces the following
f-structure and meaning constructor:

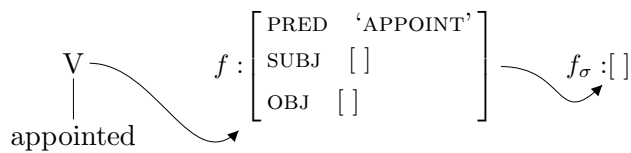
\vbox{\enumsentence{($f$ \pred) = `{\sc appoint}'\\[2em]
\modsmalltree{1}{\node{l}{V}\\
            \node{m}{appointed}}
\hspace*{2em}%
\fd{$f:$\node{a}{\fdand{
       \feat{\pred}{`{\sc appoint}'}
       \feat{\subj}{[\ ]}
       \feat{\obj}{[\ ]}}}}
\hspace*{2em}%
\fd{$f_\sigma:$\node{b}{[\ ]}}\\[1em]
$\All{ X, Y}\attr{f}{\subj}_\sigma\means X \otimes
\attr{f}{\obj}_\sigma\means Y \linimp f_\sigma \means \IT{appoint}\/(X, Y)$
\anodecurve[r]{l}[bl]{a}{2.5em}
\anodecurve[r]{a}[bl]{b}{2.5em}
\nodeconnect{l}{m}}}

\noindent The instantiated meaning constructor asserts that
$f$ is the f-structure for a clause with predicate ($\pred$) `{\sc
appoint}', and:
\begin{itemize}
\item if $f$'s subject $\attr{f}{\subj}$ has
meaning $X$
\item and ($\otimes$) $f$'s object $\attr{f}{\obj}$ has meaning $Y$
\item then ($\linimp$) $f$ has meaning $\IT{appoint}(X,Y)$.
\end{itemize}

It is not an accident that the form of the meaning constructor for
\IT{appointed} is analogous to the type $(e\times e) \rightarrow t$
which, in its curried form \mbox{$e\rightarrow{e}\rightarrow{t}$}, is
the standard type for a transitive verb in a compositional semantics
setting \cite{LTFGamut:vol2}. In general, the propositional structure
of the meaning constructors of lexical entries will parallel the types
assigned to the meanings of the same words in compositional analyses.

As mentioned above, in most cases phrase-structure rules make no
semantic contributions of their own. Thus, all the semantic
information for a sentence like `Bill appointed Hillary' is provided
by the lexical entries for `Bill', `appointed', and `Hillary':

\enumsentence{
\lexentry{Bill}{NP}{
\attr{\Up}{\pred} = `{\sc Bill}'\\ $\Ups \means \IT{Bill}$}

\lexentry{appointed}{V}{
\attr{\Up}{\pred}= `{\sc appoint}'\\
$\All{ X, Y}\attr{\Up}{\subj}_\sigma\means X \otimes
\attr{\Up}{\obj}_\sigma\means Y \linimp\Ups \means \IT{appoint}\/(X, Y)$}

\lexentry{Hillary}{NP}{
\attr{\Up}{\pred} = `{\sc Hillary}'\\
$\Ups \means \IT{Hillary}$}\label{ex:bah-lex}}

\paragraph{Assembly of meanings via deduction}
We have now the ingredients for building semantic interpretations by
deductive means. To recapitulate the development so far, lexical
entries provide semantic constructors, which are linear-logic formulas
specifying how the meanings of f-structures are built from the
meanings of their substructures. Thus, linear logic serves as a {\em
glue language} to assemble meanings. Certain terms in the glue
language represent (open) formulas of an appropriate {\em meaning
language}, which for the present purposes will be a version of
Montague's intensional logic \cite{Montague:PTQ}.\footnote{The reader
familiar with Montague may be surprised by the apparently purely
extensional form of the meaning terms in the examples that follow, in
contrast with Montague's use of intensional expressions even in
purely extensional cases to allow for uniform translation rules. The
reasons for this divergence are explained in Section
\ref{sec:intension}.} Other terms in the glue language represent
semantic projections. The glue-language formula $f \means t$, with $f$
a term representing a semantic projection and $t$ a term representing
a meaning-language formula, expresses the association between the
semantic projection denoted by $f$ and the meaning fragment denoted by
$t$.

The fragment of linear logic we use as glue language will be described
incrementally as we discuss examples, and is summarized in Appendix
\ref{sec:syn-app}. The semantic contribution of each lexical entry is a
linear-logic formula, its meaning constructor, that can be understood
as ``instructions'' for combining the meanings of the lexical entry's
syntactic arguments to obtain the meaning of the f-structure headed by
the entry.  In the case of the verb `appointed' above, the meaning
constructor is a glue language formula consisting of instructions on
how to assemble the meaning of a sentence with main verb `appointed',
given the meanings of its subject and object.

We will now show how meanings are assembled by linear-logic deduction.
The full set of proof rules relevant to this paper is given in
Appendix \ref{sec:rules-app}. For readability, however, we will
present derivations informally in the main body of the paper. As a
first example, consider the lexical entries in \pex{ex:bah-lex} and
let the constants $f$, $g$ and $h$ name the following f-structures:

\enumsentence{\label{ex:bahafs}\evnup{\fd{
$f$:\fdand{\feat{\pred}{`{\sc appoint}'}
       \feat{\subj}{$g$:\fdand{\feat{\pred}{`{\sc Bill}'}}}
       \feat{\obj}{$h$:\fdand{\feat{\pred}{`{\sc Hillary}'}}}}}}}

\noindent Instantiating the lexical entries for `Bill',
`Hillary', and `appointed' appropriately, we obtain the following
meaning constructors, abbreviated as \BF{bill}, \BF{hillary}, and
\BF{appointed}:
\[
\begin{array}{@{\strut}ll@{\strut}}
\BF{bill}\colon& g_{\sigma} \means \IT{Bill}\\
\BF{hillary}\colon& h_{\sigma} \means \IT{Hillary}\\
\BF{appointed}\colon& \All{ X, Y} g_{\sigma}\means X \otimes
h_{\sigma}\means Y \linimp f_{\sigma}\means \IT{appoint}(X, Y)
\end{array}
\]

\noindent These formulas show how the generic semantic contributions in
the lexical entries are instantiated to reflect their participation in
this particular f-structure.  Since the entry `Bill' gives rise to
f-structure $g$, the meaning constructor for `Bill' provides a meaning
for $g_{\sigma}$. Similarly, the meaning constructor for `Hillary'
provides a meaning for $h_\sigma$.  The verb `appointed' requires two
pieces of information, the meanings of its subject and object, in no
particular order, to produce a meaning for the clause.  As
instantiated, the f-structures corresponding to the subject and object
of the verb are $g$ and $h$, respectively, and $f$ is the f-structure
for the entire clause.  Thus, the instantiated entry for `appointed'
shows how to combine a meaning for $g_{\sigma}$ (its subject) and
$h_{\sigma}$ (its object) to generate a meaning for $f_{\sigma}$ (the
entire clause).

In the following, assume that the formula \BF{bill-appointed} is
defined thus:
\[\begin{array}[t]{ll}
\BF{bill-appointed}\colon&
\All{ Y}h_{\sigma}\means Y \linimp f_{\sigma} \means \IT{appoint}(\IT{Bill}, Y)
\end{array}
\]
\noindent Then the following derivation is possible in linear logic
($\vdash$ stands for the linear-logic entailment relation):
\enumsentence{\label{ex:bahderiv}
$
\begin{array}[t]{l@{\hspace*{2em}}ll}
&\BF{bill} \otimes \BF{hillary} \otimes \BF{appointed} & (Premises.) \\[.5ex]
\vdash &  \BF{bill-appointed} \otimes \BF{hillary} &
X\mapsto \IT{Bill}\\[0.5ex]
\vdash & f_{\sigma} \means \IT{appoint}(\IT{Bill}, \IT{Hillary}) &
Y\mapsto \IT{Hillary}
\end{array}
$}
Each formula is annotated with the variable substitutions (universal
instantiations) required to derive it from the preceding one by the
modus ponens rule $A\otimes (A\linimp B) \vdash B$.

Of course, another derivation is also possible. Assume that the
formula \BF{appointed-hillary} is defined as:
\[\begin{array}[t]{ll}
\BF{appointed-hillary}\colon&
\All{ X}g_{\sigma}\means X \linimp f_{\sigma} \means
\IT{appoint}(X, \IT{Hillary})
\end{array}
\]
\noindent Then we have the following derivation:
\enumsentence{
$
\begin{array}[t]{l@{\hspace*{2em}}ll}
&\BF{bill} \otimes \BF{hillary} \otimes \BF{appointed} & (Premises.) \\[.5ex]
\vdash &  \BF{bill} \otimes \BF{appointed-hillary} &
Y\mapsto \IT{Hillary}\\[0.5ex]
\vdash & f_{\sigma} \means \IT{appoint}(\IT{Bill}, \IT{Hillary}) &
X \mapsto \IT{Bill}
\end{array}
$}

In summary, each word in a sentence contributes a linear-logic
formula, its meaning constructor, relating the semantic projections of
specific f-structures in the LFG analysis to representations of their
meanings.  From these glue language formulas, the interpretation
process attempts to deduce an atomic formula relating the semantic
projection of the whole sentence to a representation of the sentence's
meaning. Alternative derivations may yield different such conclusions,
corresponding to ambiguities of semantic interpretation.

\paragraph{Linear logic} As we have just outlined, we
use deduction in linear logic to assign meanings to sentences,
starting from information about their functional structure and about
the semantic contributions of their words. Traditional compositional
approaches depend on a strict separation between functors and
arguments, typically derived from a binary-branching phrase-structure
tree. In contrast, our linear-logic-based approach allows the premises
carrying semantic information to commute while keeping their
connection to the f-structure, and is thus more compatible with the
flat and relatively free form organization of functional structure.

An important motivation for using linear logic is that it allows us to
directly capture the intuition that lexical items and phrases each
contribute exactly once to the meaning of a sentence.  As noted by
\namecite[page~172]{KleinSag:Type}:
\begin{quote}
Translation rules in Montague semantics have the property that the
translation of each component of a complex expression occurs exactly
once in the translation of the whole.  \ldots That is to say, we do
not want the set S [of semantic representations of a phrase] to
contain {\em all\/} meaningful expressions of IL which can be built up
from the elements of S, but only those which use each element exactly
once.
\end{quote}

\noindent In our terms, the semantic contributions of the constituents
of a sentence are not context-independent assertions that may be used
or not in the derivation of the meaning of the sentence depending on
the course of the derivation. Instead, the semantic contributions are
{\em occurrences\/} of information which are generated and used exactly
once.  For example, the formula $g_{\sigma}\means
\IT{Bill}$ can be thought of as providing one occurrence of the meaning
$\IT{Bill}$ associated to the semantic projection $g_{\sigma}$.  That
meaning must be consumed exactly once (for example, by \BF{appointed} in
\pex{ex:bahderiv}) in the derivation of a meaning of the entire utterance.

It is this ``resource-sensitivity'' of natural language semantics---an
expression is used exactly once in a semantic derivation---that linear
logic can model. The basic insight underlying linear logic is that
logical formulas are {\em resources\/} that are produced and consumed in
the deduction process.  This gives rise to a resource-sensitive notion
of implication, the {\em linear implication\/} $\linimp$: the formula $A
\linimp B$ can be thought of as an action that can {\em consume\/} (one
copy of) $A$ to produce (one copy of) $B$. Thus, the formula $A
\otimes (A \linimp B)$ linearly entails $B$.  It does not entail $A
\otimes B$ (because the deduction consumes $A$), and it does not entail
$(A \linimp B) \otimes B$ (because the linear implication is also
consumed in doing the deduction).  This resource-sensitivity not only
disallows arbitrary duplication of formulas, but also disallows
arbitrary deletion of formulas. Thus the linear multiplicative
conjunction $\otimes$ is sensitive to the multiplicity of formulas: $A
\otimes A$ is not equivalent to $A$ (the former has two copies of the
formula $A$).  For example, the formula $A \otimes A \otimes (A
\linimp B)$  linearly entails $A \otimes B$ (there is still one $A$
left over) but does not entail $B$ (there must still be one
$A$ present).  In this way, linear logic checks that a formula is used
once and only once in a deduction, enforcing the requirement that
each component of an utterance contributes exactly once to the
assembly of the utterance's meaning.

A direct consequence of the above properties of linear logic is that
the constraints of functional {\em completeness}\/ and {\em coherence}
hold without further stipulation\footnote{`An f-structure is {\it
locally complete}\/ if and only if it contains all the governable
grammatical functions that its predicate governs.  An f-structure is
{\em complete}\/ if and only if all its subsidiary f-structures are
locally complete. An f-structure is {\em locally coherent}\/ if and
only if all the governable grammatical functions that it contains are
governed by a local predicate.  An f-structure is {\em coherent}\/ if
and only if all its subsidiary f-structures are locally coherent.'
\cite[pages~211--212]{KaplanBresnan:LFG}

To illustrate:
\enumsentence[(a)]{*John devoured. [incomplete]}
\enumsentence[(b)]{*John arrived Bill the sink. [incoherent]}}
\cite{DLS:EACL}.  In the present setting, the feature structure $f$
corresponding to the utterance is associated with the ($\otimes$)
conjunction $\phi$ of all the formulas associated with the lexical
items in the utterance. The conjunction is said to be {\em complete\/}
and {\em coherent\/} iff $Th \vdash \phi \linimp f_{\sigma} \means t$ (for
some term $t$), where $Th$ is the background theory of general
linguistic principles.  Each $t$ is to be thought of as a valid
meaning for the sentence. This guarantees that the entries are used
exactly once in building up the denotation of the utterance: no
syntactic or semantic requirements may be left unfulfilled, and no
meaning may remain unused.

Our glue language needs to be only a fragment of higher-order linear
logic, the {\em tensor fragment}, that is closed under conjunction,
universal quantification, and implication.  This fragment arises from
transferring to linear logic the ideas underlying the concurrent
constraint programming scheme of \namecite{Saraswat:PhD}.\footnote{
\namecite{SaraswatLincoln:HLCC} provide
an explicit formulation for the higher-order version of the linear
concurrent constraint programming scheme. \namecite{Scedrov:Linear}
gives a tutorial introduction to linear logic itself;
\namecite{Saraswat:IntroLCC} supplies further background on
computational aspects of linear logic relevant to the implementation
of the present proposal.}

\paragraph{Relationship with Categorial Syntax and Semantics}

As suggested above, there are close connections between our
approach and various systems of categorial syntax and semantics. The
Lambek calculus \cite{Lambek:SentStruct}, introduced as a logic of
syntactic combination, turns out to be a fragment of noncommutative
multiplicative linear logic.  If permutation is added to Lambek's
system, its left- and right-implication connectives ($\setminus$ and
$/$) collapse into a single implication connective with behavior
identical to $\linimp$. This undirected version of the Lambek calculus
was developed by van Benthem
\shortcite{vanBenthem:lambek,VanBenthem:LgInAction} to account for the
semantic combination possibilities of phrase meanings.

Those systems and related ones
\cite{Moortgat:categorial,Hepple:PhD,Morrill:intensional} were
developed as calculi of syntactic/semantic types, with propositional
formulas representing syntactic categories or semantic types. Given
the types for the lexical items in a sentence as assumptions, the
sentence is syntactically well-formed in the Lambek calculus if the
type of the sentence can be derived from the assumptions arranged as
an ordered list. Furthermore, the Curry-Howard isomorphism between
proofs and terms \cite{Howard:construction} allows the extraction of a
term representing the meaning of the sentence from the proof that the
sentence is well-formed \cite{vanBenthem:cat-lambda}. However, the
Lambek calculus and its variants carry with them a particular view of
the syntax-semantics interface which is not obviously compatible with
the flatter f-structures of LFG.  In Section \ref{sec:categorial}, we
will examine more closely the differences between those approaches and
ours.

On the other hand, categorial semantics in the undirected Lambek
calculus and other related commutative calculi provide an analysis of
the possibilities of meaning combination independently of the
syntactic realizations of those meanings, but does not offer a
mechanism for relating semantic combination possibilities to the
corresponding syntactic combination possibilities.

Our system follows categorial semantics in using the ``propositional
skeleton'' of glue formulas to encode the types of phrase meanings and
thus their composition potential. In addition, however, first-order
quantification over semantic projections maintains the connection
between those types and the corresponding syntactic objects, while
quantification over semantic terms is used to build the meanings of
those syntactic objects. This tripartite organization reflects the
three linked systems of representation that participate in semantic
interpretation: syntactic structure, semantic types and semantic
interpretations themselves.  In this way, we can take advantage of the
principled description of potential meaning combinations arising from
categorial semantics without losing track of the constraints imposed
by syntax on the possible combinations of those meanings.

\section{Quantification}

Our treatment of quantification, and in particular of quantifier scope
ambiguities and of the interactions between scope and bound anaphora,
follows the analysis of Pereira
\shortcite{Pereira:SemComp,Pereira:HOD}, but offers in addition a formal
account of the syntax-semantics interface, which was treated only
informally in that earlier work.

\subsection{Quantifier meanings} The basic idea for the analysis can be
seen as a logical counterpart at the glue level of the standard type
assignment for generalized quantifiers
\cite{Barwise+Cooper:generalized}. The generalized quantifier
meaning of a natural language determiner has the following type:
\enumsentence{$(e\rightarrow
t)\rightarrow (e\rightarrow t) \rightarrow t$\label{eq:gen-quant-type}}
\noindent that is, the type of
functions from two properties, the quantifier's restriction and scope,
to propositions. At the semantic glue level, we can understand
that type as follows. For any determiner, if for arbitrary $x$ we can
construct a meaning $R(x)$ for the quantifier's restriction, and again
for arbitrary $x$ we can construct a meaning $S(x)$ for the
quantifier's scope, where $R$ and $S$ are suitable properties (functions from
entities to propositions), then we can construct the meaning $Q(R,S)$
for the whole sentence containing the determiner, where $Q$ is the
meaning of the determiner.

Assume for the moment that we have determined the following semantic
structures: $\IT{restr}$ for the restriction (a common noun phrase),
$\IT{restr-arg}$ for its implicit argument, $\IT{scope}$ for the scope
of quantification, and $\IT{scope-arg}$ for the grammatical function
filled by the quantified NP.  Then the foregoing analysis can be
represented in linear logic by the following schematic
formula:\footnote{We use lower-case letters for {\em essentially
universal\/} variables, that is, variables that stand for new local
constants in a proof. We use capital letters for {\em essentially
existential\/} variables, that is, Prolog-like variables that become
instantiated to particular terms in a proof. In other words,
essentially existential variables stand for specific but as yet
unspecified terms, while essentially universal variables stand for
arbitrary constants, that is, constants that could be replaced by {\em
any} term while still maintaining the validity of the derivation. In
the linear-logic fragment we use here, essentially existential
variables arise from universal quantification with outermost scope,
while essentially universal variables arise from universal
quantification whose scope is a conjunct in the antecedent of an
outermost implication.}

\enumsentence{$\begin{array}[t]{r@{\,}l}
\All{ R, S} & (\All{ x}\IT{restr-arg} \means x \linimp \IT{restr}
\means R(x))\\
\otimes & (\All{ x}\IT{scope-arg} \means x \linimp \IT{scope} \means S(x))\\
\linimp & \IT{scope} \means Q(R, S)
\end{array}$\label{eq:gen-quant-lin}}
Given the equivalence between $A\otimes B\linimp C$ and $A\linimp
(B\linimp C)$, the propositional part of \pex{eq:gen-quant-lin}
parallels the generalized quantifier type \pex{eq:gen-quant-type}.

In addition to providing a semantic type assignment for determiners,
\pex{eq:gen-quant-lin} uses glue language quantification to express how
the meanings of the restriction and scope of quantification are
determined and combined into the meaning of the quantified clause.
The subformula
$$\All{ x}\IT{restr-arg} \means x \linimp
\IT{restr} \means R(x)$$
specifies that $\IT{restr}$ has meaning $R(x)$ if
for arbitrary $x$
$\IT{restr-arg}$ has meaning $x$,
that is, it gives the dependency of the meaning of a common noun
phrase on its implicit argument. Property $R$ is the representation of
that dependency as a function in the meaning language. Similarly, the
subformula
$$\All{x}\IT{scope-arg} \means x \linimp \IT{scope} \means
S(x)$$
specifies the dependency of the meaning $S(x)$ of a semantic
structure $\IT{scope}$ on the meaning $x$ of one of its arguments
$\IT{scope-arg}$. If both dependencies hold, then $R$ and $S$ are an
appropriate restriction and scope for the determiner meaning $Q$.

Computationally, the nested universal quantifiers substitute unique
new constants (eigenvariables) for the quantified variable $x$, and
the nested implications try to prove their consequents with their
antecedents added to the current set of assumptions. For the
restriction (the case of the scope is similar), this will in
particular involve solving an equation of the form $R(x) = t$, where
$\IT{restr}\means t$ has been derived. The equation must be solved
modulo $\alpha$-, $\beta$- and $\eta$-conversion, and any solution $R$
must not contain occurrences of $x$, since $R$'s scope is wider than
$x$'s. Higher-order unification \cite{Huet:HOU} is a procedure suitable
for solving such equations.\footnote{While higher-order unification
is in general undecidable, the unification problems involved here are
of one of the forms $F(x) = t$ or $p(X) = t$ where $t$ is a closed term,
$F$ and $X$ essentially existential variables and $x$ and $p$
essentially universal variables. These cases fall within the
$l\lambda$ fragment of \namecite{Miller:LLambda}, which is a
decidable extension of first-order unification.}

\subsection{Quantifier restrictions}
We have seen that since the meaning of the restriction of a quantifier
is a property (type $e\rightarrow t$), its meaning constructor has
the form of an implication, just like a verb.  In
\pex{eq:gen-quant-lin}, the first line of the determiner's semantic
constructor
$$(\All{ x}\IT{restr-arg} \means x \linimp \IT{restr} \means R(x))$$
requires a meaning $x$ for {\em restr-arg} to produce the meaning
$R(x)$ for {\em restr}, defining the restriction $R$ of the
quantifier. We need thus to identify the semantic projections
$\IT{restr-arg}$ and $\IT{restr}$.

The f-structure of a quantified NP has the general
form: \enumsentence{\evnup{ \fd{$f$:\fdand{\feat{\spec}{$q$}
\feat{\pred}{$n$}}}}} where $q$ is the determiner f-structure and $n$
the noun f-structure. None of the f-structures $f$, $q$ or $n$ is a
natural syntactic correlate of the argument or result of the
quantifier restriction.  This contrasts with the treatment of verbs,
whose semantic contributions and argument dependencies are directly
associated with appropriate syntactic units of the clauses they head.
Therefore, we take the semantic projection $f_\sigma$ of the
quantified NP to be structured with two attributes
$\attr{f_\sigma}{\var}$ and $\attr{f_\sigma}{\restr}$:

\vbox{\enumsentence{
\modsmalltree{1}{\node{l}{Det}\\
            \node{m}{every}}
\hspace*{2em}%
\fd{$f:$\node{a}{\fdand{
       \feat{\spec}{`{\sc every}'}}}}
\hspace*{2em}%
\fd{$f_\sigma:$\node{b}{\fdand{\feat{\sc var}{[\ ]}
                               \feat{\sc restr}{[\ ]}}}}
\anodecurve[br]{l}[bl]{a}{2em}
\anodecurve[br]{a}[bl]{b}{2em}
\nodeconnect{l}{m}}}

\noindent The value of {\sc var} will play the role of {\em restr-arg},
supplying an entity-type variable, and the value of {\sc restr} will
play the role of {\em restr} in the meaning constructor of the
determiner.  For a preliminary version of the lexical entry for
`every', we replace the relevant portions of our canonical determiner
entry appropriately:

\enumsentence{Preliminary lexical entry for `every':

\lexentry{every}{Det}{\label{ex:every-entry-prelim}
\attr{\Up}{\spec} = `{\sc every}'\\
$\begin{array}[t]{r@{\,}l}
\All{ R, S} & (\All{x} \attr{\Ups}{\var}\means x \linimp
\attr{\Ups}{\restr} \means R(x)) \\
\otimes & (\All{ x}\IT{scope-arg} \means x \linimp \IT{scope} \means S(x))\\
\linimp & \IT{scope} \means \IT{every}(R, S)
\end{array}$}}
The restriction property $R$ should of course be derived from the
semantic contribution of the nominal part of the noun
phrase. Therefore, semantic constructors for nouns must connect
appropriately to the {\sc var} and {\sc restr} components of the noun
phrase's semantic projection, as we shall now see.
\subsection{Noun meanings}
We will use the following phrase structure rule for simple noun
phrases:
\enumsentence{\label{ex:np-rule}
\phraserule{NP}{
\rulenode{Det\\ \up = \Down}
\rulenode{N\\ \up = \Down}}}
This rule states that the determiner Det and noun N contribute equally
to the f-structure for the NP.  Lexical specifications ensure that the
noun contributes the \pred\ attribute and its value, and the
determiner contributes the \spec\ attribute and its value.

The c-structure, f-structure, and semantic structure for `every
voter', together with the functional relations between them, are:

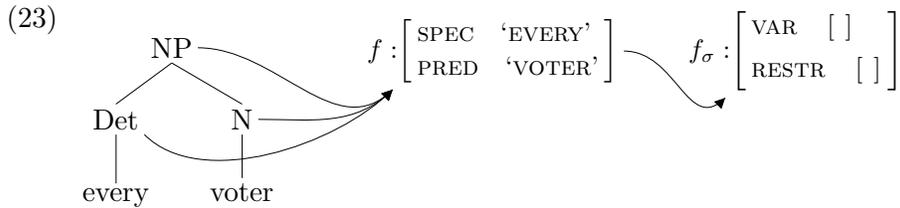
\vbox{\enumsentence{\label{eg:ev-voter}\evnup{
\modsmalltree{3}{\mc{3}{\node{l}{NP}\hspace*{1em}}\\
            \node{m}{Det}&\mc{2}{\hspace*{1em}\node{n}{N}}\\
            \node{o}{every}&\mc{2}{\hspace*{1em}\node{p}{voter}}}
\hspace*{2em}%
\fd{$f:$\node{a}{\fdand{\feat{\spec}{`{\sc every}'}
                        \feat{\pred}{`{\sc voter}'}}}}
\hspace*{2em}%
\fd{$f_\sigma:$\node{b}{\fdand{\feat{\sc var}{[\ ]}
                               \feat{\sc restr}{[\ ]}}}}
\anodecurve[r]{l}[bl]{a}{3em}
\anodecurve[br]{m}[bl]{a}{3em}
\anodecurve[r]{n}[bl]{a}{2em}
\anodecurve[r]{a}[bl]{b}{2em}
\nodeconnect{l}{m}
\nodeconnect{l}{n}
\nodeconnect{m}{o}
\nodeconnect{n}{p}}}}

\noindent In rule \pex{ex:np-rule}, the
meaning constructors of the noun `voter' and the determiner `every'
make reference to the same semantic structure, and in particular to
the same semantic projections {\sc var} and {\sc restr}.  The noun
will supply appropriate values for the {\sc var} and {\sc restr}
attributes of the NP, and these will be consumed by the
determiner's meaning constructor.  Thus, the semantic constructor
for a noun will have the general form $$\All{ x} \attr{\Ups}{\var}
\means x \linimp \attr{\Ups}{\restr} \means P x$$ where $P$ is the
meaning of the noun.\footnote{Of course, the derivation would be more
complicated if the NP included adjective phrases or other
noun modifiers; for the sake of brevity, we will not discuss the
contribution of noun modifiers in this paper.  Intuitively, the
function of modifiers is to consume the meaning of the phrase they
modify and produce a new, modified meaning of the same semantic shape,
which can play the same semantic role as the unmodified phrase can
play.  \namecite{DLS:EACL} provide a general discussion of modification in
this framework.} In particular, the lexical entry for
`voter' is:
\enumsentence{
\lexentry{voter}{N}{
\attr{\Up}{\pred} = `{\sc voter}'\\
$\All{ X}\attr{\Ups}{\var}\means X \linimp \attr{\Ups}{\restr} \means
\IT{voter}(X)$}}
Given this entry and the one for `every' in
\pex{ex:every-entry-prelim}, we obtain the following instantiated
semantic constructors for
\pex{eg:ev-voter}:
$$\begin{array}{ll}
\BF{every}\colon&
\begin{array}[t]{r@{\,}l}
\All{R, S} & (\All{ x} \attr{f_\sigma}{\var} \means x \linimp
\attr{f_\sigma}{\restr} \means R(x)) \\
\otimes & (\All{ x}\IT{scope-arg} \means x \linimp \IT{scope} \means S(x))\\
    \linimp & \IT{scope} \means \IT{every}(R, S)
\end{array}\\[2ex]
\BF{voter}\colon&
\All{ X} \attr{f_\sigma}{\var} \means X \linimp
\attr{f_\sigma}{\restr}
\means \IT{voter}(X)
\end{array}$$
\noindent Applying the variable substitutions
$X \mapsto x, R \mapsto \IT{voter}$ and modus ponens to those two
premises, we obtain the semantic constructor for
`every voter':
\enumsentence{\label{ex:every-voter-schema}
$\begin{array}[t]{lr@{\,}l}
\BF{every-voter}\colon\ &
\All{S} & (\All{x} \IT{scope-arg} \means x \linimp \IT{scope} \means S(x))\\
&\linimp & \IT{scope} \means \IT{every}(\IT{voter}, S)
\end{array}$}
In keeping with the parallel noted earlier between our semantic
constructors and compositional types, the propositional part of this
formula corresponds to the standard type for NP meanings,
$(e\rightarrow t)\rightarrow t$.

\subsection{Quantified NP meanings}
\label{sec:qnp-mean}

To complete our analysis of the semantic contribution of determiners,
we need to characterize how a quantified NP contributes to the
semantics of a sentence in which it appears, by specifying the
semantic projections $\IT{scope-arg}$ and $\IT{scope}$ in quantified
NP semantic constructors like (\ref{ex:every-voter-schema}).

\paragraph{Individual-type contribution} First, we require the meaning of
the scope to depend on the meaning of (the position filled by) the
quantifier itself.  Thus, {\em scope-arg}\/ is the semantic projection
for the quantified NP itself:

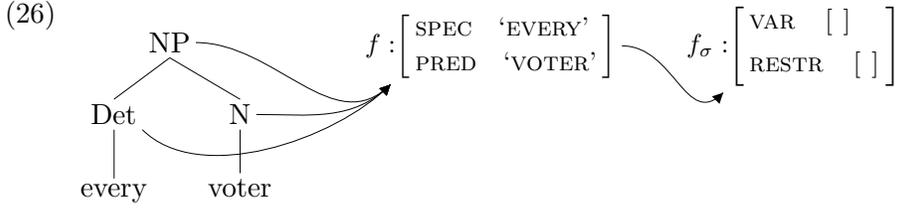
\vbox{\enumsentence{\label{ex:every-voter-unscoped}\evnup{
\modsmalltree{3}{\mc{3}{\node{l}{NP}\hspace*{1em}}\\
            \node{m}{Det}&\mc{2}{\hspace*{1em}\node{n}{N}}\\
            \node{o}{every}&\mc{2}{\hspace*{1em}\node{p}{voter}}}
\hspace*{2em}%
\fd{$f:$\node{a}{\fdand{\feat{\spec}{`{\sc every}'}
                        \feat{\pred}{`{\sc voter}'}}}}
\hspace*{2em}%
\fd{$f_\sigma:$\node{b}{\fdand{\feat{\sc var}{[\ ]}
                               \feat{\sc restr}{[\ ]}}}}
\anodecurve[r]{l}[bl]{a}{3em}
\anodecurve[br]{m}[bl]{a}{3em}
\anodecurve[r]{n}[bl]{a}{2em}
\anodecurve[r]{a}[bl]{b}{2em}
\nodeconnect{l}{m}
\nodeconnect{l}{n}
\nodeconnect{m}{o}
\nodeconnect{n}{p}}\\[1em]
$\BF{every-voter}\colon\
\All{ S}  (\All{ x}  f_\sigma \means x \linimp \IT{scope} \means S(x))
\linimp  \IT{scope} \means \IT{every}(\IT{voter}, S)$}}

\noindent Informally, the constructor for `every voter' can be read as
follows: if by giving the arbitrary meaning $x$ of type $e$ to $f$,
the f-structure for `every voter', we can derive the meaning $S(x)$ of
type $t$ for the scope of quantification {\em scope}\/, then $S$ can
be the property that the quantifier requires as its scope, yielding
the meaning $\IT{every}(\IT{voter}, S)$ for {\em scope}. The
quantified NP can thus be seen as providing two contributions
to an interpretation: locally, a {\em referential import\/} $x$, which
must be discharged when the scope of quantification is established;
and globally, a {\em quantificational import\/} of type
\mbox{$(e\rightarrow t)\rightarrow t$}, which is applied to the
meaning of the scope of quantification to obtain a quantified
proposition.

Notice also that the assignment of a meaning to {\em scope}\/ appears
on both sides of the implication, and that in fact the meaning is not
the same in the two instances.  Linear logic allows for the {\it
consumption}\/ of the preliminary meaning in the antecedent of the
implication, {\em producing} the final meaning for {\em scope}\/ in the
conclusion.

\paragraph{Scope of quantification}
To complete our account of quantified NP interpretation,
we need to explain how to select the possible scopes of
quantification, for which we used the place-holder $\IT{scope}$ in
(\ref{ex:every-voter-unscoped}).

As is well known, the scope of a quantifier is not syntactically
fixed. While syntactic effects may play a significant role in scope
preferences, most claims of scope islands (eg. May's
\shortcite{May:lf}) are defeasible given appropriate choices of
lexical items and context. Therefore, we will take as possible quantifier
scopes all semantic projections for which a meaning of proposition
type can be derived. But even this liberal notion of scope is subject
to indirect constraints from syntax, as those that we will see arise
from interaction of coreference relations and quantification.

Previous work on scope determination in LFG
\cite{HalvorsenKaplan:Projections} defined possible scopes at the
f-structure level, using {\em inside-out functional uncertainty}\/ to
nondeterministically choose a scope f-structure for quantified noun
phrases.  That approach requires the scope of a quantified NP
to be an f-structure which contains the NP f-structure.  In
contrast, our approach depends only on the logical form of
semantic constructors to yield just the appropriate scope
choices. Within the constraints imposed by that logical form, the
actual scope can be freely chosen. Logically, that means that the
semantic constructor for an NP should quantify universally
over scopes, as follows:
\enumsentence{\label{ex:every-voter}
\evnup{$\BF{every-voter}\colon\
\All{H, S}  (\All{x}  f_\sigma \means x \linimp H \means S(x))
\linimp  H \means \IT{every}(\IT{voter}, S)$}}

The foregoing argument leads to the following general semantic constructor
for a determiner with meaning $Q$:
\enumsentence{\label{ex:det-def}
$\begin{array}[t]{ll}
\All{ {H}, R, S} \\
\quad (\All{ x} \attr{\Ups}{\var} \means x& \left\{ \parbox{40ex}{if,
by assuming an arbitrary meaning $x$ for $(\Ups$ \var),} \right. \\[2.5ex]
\qquad \linimp \attr{\Ups}{\restr} \means R(x))& \left\{ \parbox{40ex}{a
meaning $R(x)$ for $(\Ups$ \restr) can be derived,} \right. \\[2.5ex]
\quad (\All{x}\Ups \means x & \left\{ \parbox{40ex}{and if, by
assuming an arbitrary meaning $x$ for $\Up$,} \right. \\[2.5ex]
\qquad \linimp H \means_t S(x)) &
\left\{ \parbox{40ex}{a meaning $S(x)$ for some scope $H$ can
be derived,} \right. \\[2.5ex]
\quad\ \linimp {H} \means_t Q(R, S) & \left\{
\parbox{40ex}{then we can derive a possible complete meaning for $H$}\right.
\end{array}$}
where ${H}$ ranges over semantic structures associated with meanings
of type $t$.

Note that the $\var$ and $\restr$ components of the semantic
projection for a quantified NP in our analysis play a similar
role to the $/\!\!/$ category constructor in PTQ \cite{Montague:PTQ},
that of distinguishing syntactic configurations with identical
semantic types but different contributions to the interpretation. The
two PTQ syntactic categories $t/e$ for intransitive verb phrases and
$t/\!\!/e$ for common noun phrases correspond to the single semantic
type $e \rightarrow t$; similarly, the two conjuncts in the antecedent
of \pex{ex:det-def} correspond to the same semantic type, encoded with
a linear implication, but to two different syntactic contexts, one
relating the predication of an NP to its implicit argument and
one relating a clause to an embedded argument.

\subsection{Simple example of quantification}

Before we look at quantifier scope ambiguity and interactions between
scope and bound anaphora, we demonstrate the basic operation of our
proposed meaning constructor for quantified NPs with a singly
quantified, unambiguous sentence:
\enumsentence{\label{ex:simple-quant}
Bill convinced every voter.}  To carry out the analysis, we need a
lexical entry for `convinced':

\enumsentence{
\lexentry{convinced}{V}{
\attr{\Up}{\pred}= `{\sc convince}'\\
$\All{ X, Y}\attr{\Up}{\subj}_\sigma\means X \otimes
    \attr{\Up}{\obj}_\sigma\means Y \linimp
    \Ups \means \IT{convince}(X, Y)$}}

\noindent The f-structure for \pex{ex:simple-quant} is:
\enumsentence{\evnup{
\fd{$f$:\fdand{\feat{\pred}{`{\sc convince}'}
           \feat{\subj}{$g$:\fdand{\feat{\pred}{`{\sc Bill}'}}}
           \feat{\obj}{$h$:\fdand{\feat{\spec}{`{\sc every}'}
                                     \feat{\pred}{`{\sc voter}'}}}}}}}

\noindent The premises for the derivation are appropriately
instantiated  meaning constructors
for `Bill' and `convinced' together with the instantiated meaning constructor
derived earlier for the quantified NP `every voter':
$$\begin{array}{ll@{\,}l}
\BF{bill}\colon & \makebox[1em][l]{$g_{\sigma} \means Bill$}\\[1ex]
\BF{convinced}\colon & \All{ X, Y} & g_{\sigma} \means X
\otimes h_{\sigma} \means Y\linimp f_{\sigma} \means convince\/(X, Y)\\[1ex]
\BF{every-voter}\colon\ & \All{H,S} &
\/(\All{x}  h_{\sigma} \means x \linimp H \means_t S(x))\\
 & \hfill \linimp & H \means_t \IT{every}(\IT{voter}, S)
\end{array}$$
Giving the name \BF{bill-convinced} to the formula
\[\begin{array}[t]{ll}
\BF{bill-convinced}\colon& \All{Y} h_{\sigma} \means Y \linimp
f_{\sigma} \means \IT{convince}(\IT{Bill}, Y) \\
\end{array}
\]
we have the derivation:
\[
\begin{array}{l@{\hspace*{1em}}ll}
&\BF{bill} \otimes \BF{convinced} \otimes \BF{every-voter} &
\mbox{\/(Premises.)} \\[0.5ex]
\vdash & \BF{bill-convinced} \otimes \BF{every-voter} & X \mapsto
\IT{Bill}\\[0.5ex]
\vdash & f_{\sigma} \means \IT{every}(\IT{voter},
\lam{z}\IT{convince}(\IT{Bill}, z)) & H \mapsto f_{\sigma}, Y  \mapsto x \\
& & S \mapsto \lam{z}\IT{convince}(\IT{Bill}, z)
\end{array}
\]
No derivation of a different formula $f_{\sigma} \means_t P$ is
possible.  The formula \BF{bill-convinced} represents the semantics of
the scope of the determiner `every'. The derivable formula \[\All{
Y}h_{\sigma} \meansub{e} Y \linimp h_{\sigma} \meansub{e} Y\] could at
first sight be considered another possible, but erroneous,
scope. However, the type subscripting of the \means\ relation used in
the determiner lexical entry requires the scope to represent a
dependency of a proposition on an individual, while this formula
represents the dependency of an individual on an individual
(itself). Therefore, it does not provide a valid scope for the
quantifier.

\subsection{Quantifier scope ambiguities}
When a sentence contains more than one quantifier, scope ambiguities
are of course possible. In our system, those ambiguities will appear
as alternative successful derivations. We will take as our example
this sentence:\footnote{To allow for apparent scope
ambiguities, we adopt a scoping analysis of indefinites, as proposed,
for example, by \namecite{Neale:Descriptions}.}
\enumsentence{\label{ae}
Every candidate appointed a manager.}
\noindent We need the following additional lexical entries:

\enumsentence{
\lexentry{a}{Det}{
\attr{\Up}{\spec} = `{\sc a}'\\
$\begin{array}[t]{@{\strut}r@{\,}l@{\strut}}
\All{ {H}, R, S} & (\All{ x}\attr{\Ups}{\var}\means x \linimp
\attr{\Ups}{\restr} \means R(x))  \\
\otimes & (\All{ x} \Ups \means x \linimp {H} \means S(x))  \\
\linimp & {H} \means \IT{a}(R, S)
\end{array}$}}

\enumsentence{
\lexentry{candidate}{N}{
\attr{\Up}{\pred} = `{\sc candidate}'\\
$\All{ X} \attr{\Ups}{\var} \means X \linimp \attr{\Ups}{\restr}
\means\IT{candidate}\/(X)$}}

\enumsentence{
\lexentry{manager}{N}{
\attr{\Up}{\pred} = `{\sc manager}'\\
$\All{ X}\attr{\Ups}{\var}\means X \linimp \attr{\Ups}{\restr}
\means\IT{manager}\/(X)$}}
The f-structure for sentence (\ref{ae}) is:
\enumsentence{\evnup{
\fd{$f$:\fdand{\feat{\pred}{`{\sc appoint}'}
           \feat{\subj}{$g$:\fdand{\feat{\spec}{`{\sc every}'}
                                     \feat{\pred}{`{\sc candidate}'}}}
           \feat{\obj}{$h$:\fdand{\feat{\spec}{`{\sc a}'}
                                     \feat{\pred}{`{\sc manager}'}}}}}}}

\noindent We can derive meaning constructors
for `every candidate' and `a manager' in the way shown in Section
\ref{sec:qnp-mean}.  Further derivations proceed from those
contributions together with the contribution of `appointed':
$$\begin{array}{lr@{\,}l}
\BF{every-candidate}\colon&
\All{G, R} &(\All{x}  g_{\sigma}\means x \linimp G \means R(x))\\
& \linimp & G \means \IT{every}(\IT{candidate}, R)\\[1ex]
\BF{a-manager}\colon& \All{H,S} & (\All{y}  h_{\sigma}
\means y\linimp H \means S(y))\\
& \linimp & H \means \IT{a}(\IT{manager}, S)\\[1ex]
\BF{appointed}\colon& \All{X, Y} & g_{\sigma} \means X\otimes
h_{\sigma} \means Y \linimp f_{\sigma} \means \IT{appoint}(X, Y)
\end{array}
$$

\noindent As of yet, we have not made any commitment about the scopes
of the quantifiers; the scope and scope meaning variables in
\BF{every-candidate} and \BF{a-manager} have not been instantiated.
Scope ambiguities are manifested in two different ways in our system:
through the choice of different semantic structures $G$ and $H$,
corresponding to different scopes for the quantified NPs,
or through different relative orders of application for quantifiers that
scope at the same point.  For this example, the second case is
relevant, and we must now make a choice to proceed. The two possible
choices correspond to two equivalent rewritings of {\bf appointed}:
\[
\begin{array}{ll}
\BF{appointed}_1\colon & \All{ X} g_{\sigma} \means X \linimp (\All{
Y}h_{\sigma} \means Y \linimp f_{\sigma} \means \IT{appoint}(X, Y)) \\
\BF{appointed}_2\colon & \All{ Y}h_{\sigma} \means Y\linimp (\All{ X}
g_{\sigma} \means X \linimp f_{\sigma} \means \IT{appoint}(X, Y))
\end{array}
\]
\noindent These two equivalent forms correspond to the two possible
ways of ``currying'' a two-argument function  $f: \alpha\times
\beta\rightarrow \gamma$ as one-argument functions: $$\lambda u.\lambda
v.f(u,v): \alpha \rightarrow (\beta \rightarrow \gamma)$$
$$\lambda v.\lambda
u.f(u,v): \beta \rightarrow (\alpha \rightarrow \gamma)$$
We select `a manager' to take narrower scope by using the
variable instantiations
\[H\mapsto f_\sigma, Y\mapsto y, S\mapsto \lam{v}\IT{appoint}(X,v)\]
and transitivity of implication to combine
\BF{appointed}$_1$ with {\bf a-manager} into:
$$\begin{array}{@{\strut}lr@{\,}l@{\strut}}
\BF{appointed-a-manager}\colon&
 \All{X} & g_{\sigma}\means X \\
& \linimp & f_{\sigma} \meansub{t} a(\IT{manager}, \lam{v}\IT{appoint}(X, v))
\end{array}$$
We have thus the derivation
\[
\begin{array}{l@{\hspace*{2em}}l}
& \BF{every-candidate} \otimes \BF{appointed}_1 \otimes \BF{a-manager}\\[0.5ex]
\vdash &  \BF{every-candidate} \otimes \BF{appointed-a-manager}\\[0.5ex]
\vdash & f_{\sigma} \meansub{t}\IT{every}(\IT{candidate},
\lam{u}
\IT{a}(\IT{manager},\lam{v}\IT{appoint}(u, v)))
\end{array}
\]
of the $\forall\exists$ reading of \pex{ae}, where the last step
uses the substitutions
\[G\mapsto f_{\sigma}, X\mapsto x, R \mapsto \lam{u}
\IT{a}(\IT{manager},\lam{v}\IT{appoint}(u, v))\]

Alternatively, we could have chosen `every candidate' to take narrow
scope, by combining \BF{appointed}$_2$ with {\bf every-candidate} to
produce:
$$\begin{array}{@{\strut}lr@{\,}l@{\strut}}
\BF{every-candidate-appointed}\colon&
 \All{Y} & h_{\sigma}\means Y \\
& \linimp & f_{\sigma} \meansub{t} \IT{every}(\IT{candidate},
\lam{u}\IT{appoint}(u, Y))
\end{array}$$
\noindent This gives the derivation
$$
\begin{array}{l@{\hspace*{2em}}l}
& \BF{every-candidate} \otimes \BF{appointed}_2 \otimes \BF{a-manager}\\[0.5ex]
\vdash &  \BF{every-candidate-appointed} \otimes \BF{a-manager}\\[0.5ex]
\vdash & f_{\sigma} \meansub{t}\IT{a}(\IT{manager},
\lam{v}\IT{every}\/(\IT{candidate},\lam{u}\IT{appoint}(u, v)))
\end{array}
$$ for the $\exists\forall$ reading. These are the only two possible
outcomes of the derivation of a meaning for \pex{ae}, as required.

\subsection{Constraints on quantifier scoping}
\label{sec:constraints}

Sentence (\ref{ex:admirer}) contains two quantifiers and therefore might
be expected to show a two-way ambiguity analogous to the one described
in the previous section:

\enumsentence{\label{ex:admirer}
Every candidate appointed an admirer of his.}

\noindent However, no such ambiguity is found if the pronoun `his'
is taken to corefer with the subject `every candidate'. In this
case, only one reading is available, in which `an admirer of his'
takes narrow scope.  Intuitively, this NP may not take wider
scope than the quantifier `every candidate', on which its
restriction depends.

As we will soon see, the lack of a wide scope `a' reading follows
automatically from our formulation of the meaning constructors for
quantifiers and anaphors without further stipulation. In Pereira's
earlier work on deductive interpretation (Pereira 1990, 1991),
\nocite{Pereira:SemComp,Pereira:HOD} the same result was achieved
through constraints on the relative scopes of glue-level universal
quantifiers representing the dependencies between meanings of clauses
and the meanings of their arguments. Here, although universal
quantifiers are used to support the extraction of properties
representing the meanings of the restriction and scope (the variables
$R$ and $S$ in the semantic constructors for determiners), the
blocking of the unwanted reading follows from the propositional
structure of the glue formulas, specifically the nested linear
implications. This is more satisfactory, since it does not reduce the
problem of proper quantifier scoping in the object language to the
same problem in the metalanguage.

The lexical entry for `admirer' is:
\enumsentence{
\lexentry{admirer}{N}{
\attr{\Up}{\pred} = `{\sc admirer}'\\
$\begin{array}[t]{r@{\,}l}
\All{ X, Y} & \attr{\Ups}{\var}\means X
\otimes \attr{\Up}{\mbox{\obl\downlett{\rm OF}}}_\sigma \means Y \\
\linimp & \attr{\Ups}{\restr} \means\IT{admirer}(X, Y)
\end{array}$}}

\noindent Here, `admirer' is a relational noun
taking as its oblique argument a phrase with prepositional marker
`of', as indicated in the f-structure by the attribute
\obl\downlett{OF}.  The meaning constructor for a relational noun
has, as expected, the same propositional form as the binary relation
type $e\times e\rightarrow t$: one argument is the admirer, and the
other is the admiree.

We assume that the semantic projection for the antecedent of the
pronoun `his' has been determined by some separate mechanism and
recorded as the $\ant$ attribute of the pronoun's semantic
projection.\footnote{The determination of appropriate values for
$\ant$ requires a more detailed analysis of other linguistic
constraints on anaphora resolution, which would need further
projections to give information about, for example, discourse
relations and salience.
\namecite{Dalrymple:SyntaxAnaph} discusses in detail LFG analyses of
anaphoric binding.} The meaning constructor of the pronoun is, then,
a formula that consumes the meaning of its antecedent and then
reintroduces that meaning, simultaneously assigning it to its own
semantic projection:
\enumsentence{\label{ex:pronoun-lex-entry}
\lexentry{his}{N}{
\attr{\Up}{\pred} = `{\sc pro}'\\ $\All{ X} \attr{\Ups}{\ant} \means X
\linimp  \attr{\Ups}{\ant} \means X
\otimes \Ups \means X$}}
\noindent In other words, the semantic contribution of a pronoun
copies the meaning $X$ of its antecedent as the meaning of the pronoun
itself.  Since the left-hand side of the linear implication
``consumes'' the antecedent meaning, it must be reinstated in the
consequent of the implication.

The f-structure for example (\ref{ex:admirer}) is:

\enumsentence{\evnup{
\fd{$f$:\fdand{\feat{\pred}{`{\sc appointed}'}
           \feat{\subj}{$g$:\fdand{\feat{\spec}{`{\sc every}'}
                                     \feat{\pred}{`{\sc candidate}'}}}
           \feat{\obj}{$h$:\fdand{\feat{\spec}{`{\sc a}'}
           \feat{\pred}{`{\sc admirer}'}
         \feat{\obl\downlett{OF}}{$i$:\fdand{\feat{\pred}{`{\sc pro}'}}}}}}}}}
\noindent with $\attr{i_\sigma}{\ant}=g_\sigma$.

We will begin by illustrating the derivation of the meaning of `an
admirer of his', starting from the following premises:
$$\begin{array}[t]{lr@{\,}l}
\BF{a}\colon& \All{H,R,S} & (\All{x} \attr{h_\sigma}{\var}\means x
  \linimp \attr{h_\sigma}{\restr} \means R(x)) \\
& \otimes & (\All{x}  h_{\sigma} \means x \linimp H \means S(x))\\
& \linimp & H \means \IT{a}(R, S) \\[1ex]
\BF{admirer}\colon& \All{Z,X} & \attr{h_\sigma}{\var} \means Z
\otimes i_{\sigma} \means X\\
& \linimp & \attr{h_\sigma}{\restr} \means \IT{admirer}(Z, X) \\[1ex]
\BF{his}\colon& \All{ X}& g_{\sigma} \means X \linimp \/g_{\sigma}
\means X \otimes
i_{\sigma} \means X
\end{array}$$
First, we rewrite {\bf admirer} into the equivalent form
$$
\All{X} i_{\sigma} \means X  \linimp (\All{Z}\attr{h_\sigma}{\var} \means Z
\linimp \attr{h_\sigma}{\restr} \means \IT{admirer}(Z, X))
$$
\noindent We can use this formula to rewrite the second conjunct in
the consequent of {\bf his}, yielding
$$\begin{array}[t]{ll}
\lefteqn{\BF{admirer-of-his}\colon} \\
& \begin{array}[t]{ll}\lefteqn{\All{ X} g_{\sigma} \means X \linimp} \\
& g_{\sigma} \means X \otimes \\
& (\All{ Z} \attr{h_\sigma}{\var} \means Z \linimp
\attr{h_\sigma}{\restr} \means \IT{admirer}(Z, X))
\end{array}
\end{array}$$
In turn, the second conjunct in the consequent of {\bf admirer-of-his}
matches the first conjunct in the antecedent of {\bf a} given
appropriate variable substitutions, allowing us to derive
$$\begin{array}{ll}
\lefteqn{\BF{an-admirer-of-his}\colon} \\
&\begin{array}[t]{@{\strut}l@{\strut}l@{\strut}l@{\strut}}
\All{X} & g_{\sigma} \means X\linimp \\
 & g_{\sigma} \means X \otimes
 (\All{H,S} & (\All{x} h_{\sigma} \means x \linimp
 H \means S(x)) \linimp \\
 & & H \means \IT{a}(\lam{z}\IT{admirer}(z, X), S))
\end{array}
\end{array}$$
\noindent At this point the other formulas available are:
$$\begin{array}[t]{l}
   \BF{every-candidate}\colon\\
   \quad\quad   \All{H,S}  (\All{x}  g_{\sigma}
\means x \linimp H \means S(x))\\
   \quad\quad \linimp  H \means \IT{every}(\IT{candidate}, S)\\[1ex]
   \BF{appointed}\colon \\
   \quad\quad \All{Z,Y}  g_{\sigma} \means Z  \otimes h_{\sigma}
   \means Y \linimp f_{\sigma} \means \IT{appoint}(Z,Y)
\end{array}$$
We have thus the meanings of the two quantified NPs.  The
antecedent implication of \BF{every-candidate} has an atomic
conclusion and hence cannot be satisfied by
\BF{an-admirer-of-his}, which has a conjunctive conclusion.
Therefore, the only possible move is to combine \BF{appointed}
and \BF{an-admirer-of-his}. We do this by first
putting {\bf appointed} in the equivalent form
\[
\All{Z}g_{\sigma} \means Z  \linimp (\All{Y} h_{\sigma}
\means Y \linimp f_{\sigma} \means \IT{appoint}(Z,Y))
\]
\noindent After substituting $X$ for $Z$, this
can be used to rewrite the first conjunct in the consequent of {\bf
an-admirer-of-his} to derive
\[
\begin{array}[t]{@{\strut}l@{\strut}l@{\strut}}
\All{X} & g_{\sigma} \means X\linimp \\
 & (\All{Y} h_{\sigma} \means Y \linimp f_{\sigma} \means
\IT{appoint}(X,Y))\otimes \\
& (\All{H,S} (\All{x} h_{\sigma} \means x \linimp
 H \means S) \linimp
 H \means \IT{a}(\lam{z}\IT{admirer}(z, X), S))
\end{array}
\]
Applying the substitutions
\[Y\mapsto x, H\mapsto f_{\sigma}, S\mapsto \lam{z}\IT{appoint}(X,z) \]
and modus
ponens with the two conjuncts in the consequent as premises, we obtain
\[ \All{X} g_{\sigma} \means X \linimp f_{\sigma}\meansub{t}
\IT{a}(\lam{z}\IT{admirer}(z,X), \lam{z}\IT{appoint}(X,z))
\]
Finally, this formula can be combined with \BF{every-candidate}
to give the meaning of the whole sentence:
\[
f_{\sigma} \meansub{t} \IT{every}(\IT{candidate}, \lam{w}
\IT{a}(\lam{z}\IT{admirer}(z,w),\lam{z}\IT{appoint}(w,z)))
\]
In fact, this is the only derivable conclusion, showing that our
analysis blocks those putative scopings in which variables occur
outside the scope of their binders.

\subsection{Adequacy}
We will now argue that our analysis is {\em sound\/} in that all
variables occur in the scope of their binders, and {\em complete\/} in
that all possible sound readings can be generated.

More precisely, soundness requires that all occurrences of a
meaning-level variable $x$ representing the argument positions
filled by a quantified NP or anaphors bound to the NP are within the
scope of the quantifier meaning of the NP binding $x$. As argued by
\namecite{Pereira:SemComp}, treatments of quantification based on
storage or quantifier raising either fail to guarantee soundness or
enforce it by stipulation. In contrast, deductive frameworks based on
a suitable type logic for meanings, such as those arising from
categorial semantics, achieve soundness as a by-product of the
soundness of their underlying type logics.

In the present setting, meaning terms are explicitly constructed
rather than read out from well-typing proofs using the Curry-Howard
connection between proofs and terms, but the particular form of our
glue-logic formulas follows that of typing rules closely and thus
guarantees soundness, as we will now explain.

Recall first that quantifiers can only be introduced into meaning
terms by quantified NPs with semantic contributions of the form
\enumsentence{$\All{H,S} (\All{x}  f \means x \linimp H \means S(x))
 \linimp H \means Q(S)$\label{ex:np-contrib}} where $f$ is the
semantic projection of the NP and $Q$ is the meaning of the NP.  Since
$S$ outscopes $x$, any instance of $S$ in a valid derivation will be a
meaning term of the form $\lam{z} T$, with $x$ not free in $T$. The
free occurrences of $z$ in $T$ will be precisely the positions
quantified over by $Q$. We need thus to show that $f$ and all
semantic projections coreferential $f$ have $z$ as its
interpretation. But $f$ itself is given interpretation $x$ in
\pex{ex:np-contrib}, while coreferential projections must by lexical
entry \pex{ex:pronoun-lex-entry} also have interpretation $x$.
Since $S(x) = T[z\mapsto x]$ with $x$ not free in $T$, any free
occurrence of $x$ in $S(x)$ must arise from substituting $x$ for $z$
in $T$. That is, the interpretation of $f$ and those of any other
projections which corefer with $f$ are quantified over by
$Q$ as required.

As seen in the above argument, the dependency of anaphors on their
antecedents is encoded by the propositional structure and
quantification over semantic projections of the semantic contributions
of anaphors. That encoding alone is sufficient to generate all and
only the possible derivations, but quantification over meaning terms
is needed to extract the appropriate meaning terms from the
derivations. The scope of glue language variables ranging over meaning
terms guarantees that all variables in meaning terms are properly
bound.

Turning now to completeness, we need to consider the correlations
between the deductive patterns and the functional structure.  With one
exception, the glue-logic formulas from which deduction starts respect
the functional structure of meanings in that implications that
conclude the meaning of a phrase depend on the meanings of all
immediate subphrases which can have meanings, or depend on the phrase
itself, but on nothing else.  The exception is anaphors, whose
meanings depend on that of their antecedents.  Thus, the meaning of a
phrase will, transitively, depend on the meanings of all its
subphrases that can have meanings and on the meanings of the
antecedents of its anaphoric pronouns.

Now we can consider the possible scopings of a quantified NP in terms
of phrase structure.  The linearity of the implication in the
antecedent of the NP's constructor requires the meaning of the scope
to depend on the meaning of the noun phrase and that nothing else
depend on that meaning.  But the above argument shows that this will
be true exactly of every containing phrase, unless there is a bound
anaphor not contained in the containing phrase that has the NP as its
antecedent.  So all the containing phrases that also contain all
coreferring anaphors are, indeed, candidates for scope of the
quantified NP.

It is worth noting that the quantificational structure of semantic
constructors is enough on its own to ensure soundness of the resulting
meaning terms. In particular, the nested implication form of
quantified NP constructors could be replaced by the flatter
$$\exists{x}.  g_{\sigma}\means x \otimes (\All{ {H}, S}  {H} \means S(x)
 \linimp {H} \means Q(S))$$
A quantifier  lexical entry would then look like:
$$\begin{array}{r@{\,}r@{\,}l} &\exists x. &
\attr{\Ups}{\var}\means x \\ & \otimes & \All{ H,R}
\attr{\Ups}{\restr} \means R(x) \linimp (\Up_{\sigma} \means x \otimes
\All{S} (H \means S(x))
\linimp H\means Q(z,Rz,Sz))
\end{array}$$

\noindent This formulation just asserts that there is a generic
entity, $x$, which stands for the meaning of the quantified phrase,
and also serves as the argument of the restriction.  The derivations
of the restriction and scope are then expected to consume this
information.  By avoiding nested implications, this formulation may be
computationally more desirable.

However, the logical structure of this formulation is not as
restrictive as that of (\ref{ex:det-def}), as it can allow additional
derivations where information intended for the restriction can be used
by the scope.  This cannot happen in our analyses, however, since all
the dependencies in semantic constructors respect syntactic
dependencies expressed in the f-structure.  As long as that principle
is observed, the formulation above is equivalent to
(\ref{ex:det-def}).  Despite this, we prefer to stay closer to
categorial semantics and thus capture explicitly quantifier and
anaphoric dependencies in the propositional structure. We will
therefore continue with the formulation (\ref{ex:det-def}).











\section{Intensional Verbs}
\label{sec:intension}
Following Montague \shortcite{Montague:PTQ},we will give an
intensional verb like {\em seek \/} a meaning that takes as direct
object an NP meaning intension.  Montague's method for assembling
meanings by function application forces the meanings of all
expressions of a given syntactic category to be raised to their lowest
common semantic type. In particular, every transitive verb meaning,
whether intensional or not, must take a quantified NP meaning
intension as direct-object argument.  In contrast, our approach allows
the semantic contributions of verbs to be of as low a type as
possible. Nonetheless, the uniformity of the translation process is
preserved because any required type changes are derivable within the
glue language, along lines similar to type change in the undirected
Lambek calculus \cite{vanBenthem:lambek}.

We will not represent intensional types explicitly at the glue level,
in contrast to categorial treatments of intensionality such as
Morrill's
\shortcite{Morrill:intensional,Morrill:type-logical}. Instead,
semantic constructors will correspond to the appropriate extensional
types. The Montagovian intension and extension operators $\intn$ and
$\extn$ will appear only at term level, to the right of $\leadsto$ in
our derivation formulas.  Thus, while the meaning of {\em seek \/} has
type $e \rightarrow (s\rightarrow ((e \rightarrow t) \rightarrow t))
\rightarrow t$, the corresponding semantic constructor in
\pex{ex:seek} parallels the type $e \rightarrow ((e \rightarrow t)
\rightarrow t) \rightarrow t$.

Our implicit treatment of intensional types imposes certain
constraints on the use of functional abstraction and application in
meaning terms, since $\beta$-reduction is only valid for intensional
terms if the argument is {\em intensionally closed}, that is, if the
free occurrences of the bound variable do not occur in intensional
contexts
\cite[p. 131]{LTFGamut:vol2}.  As we will see, that constraint is
verified by all the semantic terms in our semantic constructors. Thus,
in carrying out proofs we will be justified in solving for free
variables in meaning terms modulo the $\extn\intn$-elimination schema
${\extn}({\intn}P)=P$ and $\alpha$-, $\beta$- and $\eta$-conversion.

Generalized quantifier meanings in Montague grammar are given type
$(s\rightarrow e\rightarrow t)\rightarrow(s\rightarrow e\rightarrow
t)\rightarrow t$, that is, their are relations between properties.
While we maintain the propositional form of glue-level formulas
corresponding to the extensional generalized quantifier meanings
discussed earlier, the semantic terms in determiner semantic
constructors must be adapted to match the new intensionalized
generalized quantifier type:
\[\begin{array}[t]{@{\strut}r@{\,}l@{\strut}}
\All{H,R,S} & (\All{x}\attr{\Ups}{\var}\means x \linimp
\attr{\Ups}{\restr} \means (\extn R)(x))  \\
\otimes & (\All{x} \Ups \means x \linimp H \means (\extn S)(x))  \\
\linimp & {H} \means \IT{a}(R, S)
\end{array}\]
Therefore, the meaning of a sentence such as \pex{ex:admirer} will now
be written:
\[\IT{every}(\intn \IT{candidate}, \intn \lam{w}
\IT{a}(\intn \lam{z}\IT{admirer}(z,w),\intn \lam{z}\IT{appoint}(w,z)))
\]

The type-changing potential of the linear-logic formulation
allows us to give an intensional verb a single semantic constructor,
and yet have the expected {\em de
re\/}/{\em de dicto\/} ambiguities follow without further
stipulation. For example, we will see that for sentence
\enumsentence{\label{ex:bsu}Bill seeks a unicorn.}  we can derive the
two readings:
\[
\begin{array}{ll}
\mbox{{\em de dicto}\/ reading:} & \IT{seek}(\IT{Bill},
{\intn}\lam{Q}\IT{a}(\intn\IT{unicorn}, Q)) \\[.5ex]
\mbox{{\em de re}\/ reading:} & \IT{a}(\intn\IT{unicorn},
\lam{u}\IT{seek}(\IT{Bill}, {\intn}\lam{Q} (\extn Q)(u)))
\end{array}
\]

Given the foregoing analysis, the lexical entry for {\em seek}\/ is:
\enumsentence{\label{ex:seek}
\begin{tabular}[t]{l@{\hspace{2pt}}l@{}r@{}l}
seek & \multicolumn{3}{l}
       {($\uparrow$ {\sc pred}) = {\sc `seek'}}\\
     & $\All{Z,Y}$&&$(\uparrow \subj)_\sigma \means Z$\\
     &                &$\otimes$&$(\All{s,p}(\All{X}
   (\uparrow \obj)_\sigma \means X \linimp s \means (\extn p)(X))
   \linimp s \means Y(p))$\\
   &\multicolumn{3}{l}{\quad$\linimp\uparrow_\sigma
\means\IT{seek}(Z, {\intn}Y)$}
\end{tabular}}
which can be paraphrased as follows:
\[
\begin{array}[t]{ll}
\All{Z, Y}(\uparrow \subj)_\sigma \means Z \otimes &
\left\{\parbox{40ex}{The verb {\em seek}\/ requires a meaning $Z$ for its
subject and}\right. \\[2.5ex]
\begin{array}{ll@{\mbox{}}l}\lefteqn{(\forall s, p.} \\
&  \quad (\All{X} & (\uparrow \obj)_\sigma \means X \\
& & \linimp s \means (\extn p)(X)) \\
&  \multicolumn{2}{l}{\linimp s \means Y(p))}
\end{array}
&\left\{\parbox{40ex}{a meaning ${\intn}Y$ for its object,
where $Y$ is an NP meaning
applied to the meaning $p$ of an
arbitrarily-chosen `scope' $s$,}\right.\quad(*) \\[8ex]
\quad\linimp\uparrow_\sigma \means \IT{seek}(Z, {\intn}Y) &
\left\{\parbox{40ex}{to produce the clause meaning $\IT{seek}(Z,
{\intn}Y)$.}
\right.
\end{array}
\]
\noindent Rather than looking for an entity type meaning for its
object, the requirement expressed by the subformula labeled $(*)$
describes semantic constructors of quantified NPs. Such a constructor
takes as input the constructor for a scope, which by itself maps an
arbitrary meaning $X$ to the meaning $p(X)$ for an arbitrary scope
$s$. From that input, the quantified NP constructor will produce a
final quantified meaning $M$ for $s$. That meaning is required to
satisfy the equation $M=Y(p)$, and thus $\intn Y$ is the property of
properties (predicate intensions) that \IT{seek} requires as second
argument. Note that the argument $p$ of $Y$ in the equation will be an
intension given the new semantic constructors for
determiners. Therefore, $\beta$-conversion with the abstraction $Y$ as
functor is allowed.

The f-structure for \pex{ex:bsu} is:
\enumsentence{\evnup{
\fd{$f$:\fdand{\feat{\pred}{\sc `seek'}
           \feat{\subj}{$g$:\fdand{\feat{\pred}{\sc `Bill'}}}
           \feat{\obj}{$h$:\fdand{\feat{\spec}{\sc `a'}
                                    \feat{\pred}{\sc `unicorn'}}}}}}}
The semantic constructors associated with this f-structure are then:
$$\begin{array}[t]{l@{\hspace{2pt}}l}
\BF{seeks}\colon & \All{Z, Y}
   \begin{array}[t]{r@{}l}&g_\sigma \means Z \\
                   \otimes&(\forall s, p. (\All{X}
      h_\sigma \means X \linimp s \means (\extn p)(X))
      \linimp s \means Y(\intn p))
   \end{array} \\
   &\quad \linimp f_\sigma \means \IT{seek}(Z, {\intn}Y) \\
\BF{Bill}\colon & g_{\sigma} \means \IT{Bill}\\
\BF{a-unicorn}\colon\ & \All{H,S}
(\All{x}h_{\sigma} \means x \linimp H \means (\extn S)(x))
\linimp H \means \IT{a}(\intn\IT{unicorn}, S)
\end{array}$$
These are the premises for the deduction of the meaning of
sentence \pex{ex:bsu}.  From the premises \BF{Bill}
and \BF{seeks} and the instantiation $Z\mapsto\IT{Bill}$
we can conclude by modus ponens:
$$\begin{array}[t]{ll}
\BF{Bill-seeks}\colon & \All{Y} (\All{s, p}(\All{X}
   h_\sigma \means X \linimp s \means (\extn p)(X))
   \linimp s \means Y(p))\\
   & \qquad \linimp {f_\sigma} \means \IT{seek}(\IT{Bill}, {\intn}Y)
\end{array}$$
Different derivations starting from the premises
\BF{Bill-seeks} and \BF{a-unicorn} will yield the alternative readings
of {\em Bill seeks a unicorn\/}, as we shall now see.

\subsection{De Dicto Reading}
The formula {\bf a-unicorn}
is exactly what is required by the antecedent
of {\bf Bill-seeks} provided that the following substitutions are
performed:
\[\eqalign{H &\mapsto s \cr
S & \mapsto p \cr
X & \mapsto x \cr
Y & \mapsto \lam{P}\IT{a}(\intn\IT{unicorn},P)\cr
}\]
We can thus conclude the desired {\em de dicto}\/ reading:
$$f_\sigma \means \IT{seek}(\IT{Bill},
{\intn}\lam{P}\IT{a}(\intn\IT{unicorn},
P)))$$

To show how the premises also support a {\em de re} reading, we
consider first the simpler case of nonquantified direct objects.

\subsection{Nonquantified Objects}

The meaning constructor for {\em seek} also allows for
nonquantified objects as arguments, without needing a special
type-raising rule.  Consider the f-structure for the sentence {\it
Bill seeks Al}:

\enumsentence{\evnup{
\fd{$f$:\fdand{\feat{\pred}{\sc `seek'}
           \feat{\subj}{$g$:\fdand{\feat{\pred}{\sc `Bill'}}}
           \feat{\obj}{$h$:\fdand{\feat{\pred}{\sc `Al'}}}}}}}

\noindent The lexical entry for {\em Al}\/ is analogous to the one for
{\em Bill}\/.  We begin with the premises {\bf Bill-seeks} and {\bf
Al}:
$$\begin{array}[t]{ll}
\BF{Bill-seeks}\colon & \forall Y. (\forall s, p. (\All{X}
   h_\sigma \means X \linimp s \means (\extn p)(X))
   \linimp s \means Y(p))\\
   & \qquad \linimp {f_\sigma} \means \IT{seek}(\IT{Bill}, {\intn}Y) \\
\BF{Al}\colon\ & h_{\sigma} \means \Al
\end{array}$$
For the derivation to proceed, {\bf Al} must
supply the NP meaning constructor that {\bf
Bill-seeks} requires.  This is possible because {\bf Al} can map a proof $\Pi$
of the meaning for $s$ from the meaning for $h$ into a meaning
for $s$, simply by supplying $h_{\sigma} \means\Al$ to $\Pi$.
\begin{figure}
$$
\oneover{h_\sigma \means \Al \vdash h_\sigma \means \Al \qquad
s \means (\extn P)(\Al) \vdash  s \means (\extn P)(\Al)}
{
\oneover{h_\sigma \means \Al, h_\sigma \means \Al
\linimp s \means (\extn P)(\Al) \vdash  s \means (\extn P)(\Al)}
{
\oneover{h_\sigma \means \Al, (\forall x. h_\sigma \means x
\linimp s \means (\extn P)(x)) \vdash  s \means (\extn P)(\Al)}
{
\oneover{h_\sigma \means \Al \vdash (\forall x. h_\sigma
\means x \linimp s \means (\extn P)(x)) \linimp  s \means
(\extn P)(\Al)}
{h_\sigma \means \Al
\vdash
\forall P. (\forall x. h_\sigma
\means x \linimp s \means (\extn P)(x)) \linimp  s \means
(\extn P)(\Al)}}}}\\
$$
\caption{Proof that {\bf Al} can function as a quantifier}
\label{Al-proof}
\end{figure}
Formally, from {\bf Al} we can prove (Figure~\ref{Al-proof}):
\enumsentence{\label{type-raised-al}$\All{P}(\All{x}h_\sigma \means x
\linimp s \means (\extn P)(x)) \linimp s \means (\extn P)(\Al)$}
This corresponds to the Montagovian type-raising of a proper name
meaning to an NP meaning, and also to the undirected Lambek calculus
derivation of the sequent \mbox{$e\Rightarrow(e\rightarrow t)\rightarrow t$}.

Formula \pex{type-raised-al} with the substitutions
\[P \mapsto p, Y \mapsto \lam{P}(\extn P)(\Al)
\]
can then be used to satisfy the
antecedent of {\bf
Bill-seeks} to yield the desired result:
$$f_\sigma \means \IT{seek}(\IT{Bill}, {\intn}\lam{P}(\extn P)(\Al))$$

It is worth contrasting the foregoing derivation with treatments of
the same issue in a $\lambda$-calculus setting.  The function
$\lam{x}\lam{P}(\extn P)(x)$ raises a term like $\Al$ to the quantified NP
form $\lam{P}(\extn P)(\Al)$, so it is easy to modify $\Al$ to make it
suitable for {\bf seek\/}.  Because a $\lambda$-term must specify
exactly how functions and arguments combine, the conversion must be
explicitly applied somewhere, either in a meaning postulate or in an
alternate definition for {\em seek\/}.  Thus, it is impossible to
write a function term that is indifferent with respect to whether its
argument is $\Al$ or $\lam{P}(\extn P)(\Al)$.

In our deductive framework, on the other hand, the exact way in which
different propositions can interact is not prescribed, although it is
constrained by their logical structure.  Thus \mbox{$h_{\sigma} \means
\Al$} can function as any logical consequence of itself, in particular
as: $$\All{S,P}(\All{x}h_\sigma \means x \linimp S \means (\extn P)(x))
\linimp S \means (\extn P)(\Al)$$ This flexibility, which is also found in
syntactic-semantic analyses based on the Lambek calculus and its
variants
\cite{Moortgat:categorial,Moortgat:Labelled,VanBenthem:LgInAction},
seems to align well with some of the type flexibility in natural
language.

\begin{figure}
$$
\oneover{
\oneover{
\oneover{
\oneover{
\oneover{
\oneover{I \means Z\vdash I \means Z \qquad
S \means (\extn P)(Z) \vdash  S \means (\extn P)(Z)}
{I \means Z, I \means Z
\linimp S \means (\extn P)(Z) \vdash  S \means (\extn P)(Z)}}
{I \means Z, (\All{x} I \means x
\linimp S \means (\extn P)(x)) \vdash  S \means (\extn P)(Z)}}
{I \means Z \vdash (\All{x} I
\means x \linimp S \means (\extn P)(x)) \linimp  S \means
(\extn P)(Z)}}
{I \means Z
\vdash
\All{S,P} (\All{x}I
\means x \linimp S \means (\extn P)(x)) \linimp  S \means
(\extn P)(Z)}}
{\vdash I \means Z \linimp
\All{S,P} (\All{x}I
\means x \linimp S \means (\extn P)(x)) \linimp  S \means
(\extn P)(Z)}}
{\vdash \All{I,Z} I \means Z \linimp
\All{S,P} (\All{x}I
\means x \linimp S \means (\extn P)(x)) \linimp  S \means
(\extn P)(Z)}
$$
\caption{General Type-Raising Theorem}
\label{type-raising-proof}
\end{figure}

\subsection{Type Raising and Quantifying In}
The derivation in Figure \ref{Al-proof} can be generalized as shown in
Figure \ref{type-raising-proof} to prove the general type-raising
theorem:
\enumsentence{\label{type-raising-theorem}
$\All{I,Z} I \means Z \linimp (\All{S,P} (\forall x. I
\means x \linimp S \means (\extn P)(x)) \linimp S\means (\extn P)(Z))$
}
This theorem can be used to raise meanings of $e$ type to
$(e\rightarrow t)\rightarrow t$ type, or, dually, to quantify into
verb argument positions. For example, with the variable instantiations
\[\eqalign{I &\mapsto h_\sigma\cr
X &\mapsto  x\cr
P &\mapsto p\cr
S &\mapsto s\cr
Y &\mapsto \lam{R}(\extn R)(Z)\cr}
\]
we can use transitivity of implication to combine
\pex{type-raising-theorem} with {\bf Bill-seeks} to derive:
\[
\BF{Bill-seeks}'\colon \All{Z}h_\sigma \means Z \linimp f_\sigma
\means \IT{seek}(\IT{Bill}, {\intn}\lam{R}(\extn R)(Z))
\]
This formula can then be combined with arguments of type $e$ to
produce a meaning for $f_\sigma$. For instance, it will take the
non-type-raised $h_\sigma \means \Al$ to yield the same result
\[f_\sigma \means\IT{seek}(\IT{Bill}, {\intn}\lam{R}(\extn R)(\Al))\]
as the combination of {\bf Bill-seeks} with the type-raised version of \BF{Al}.
In fact, $\BF{Bill-seeks}'$ corresponds to type $e\rightarrow t$, and can
thus be used as the scope of a quantifier, which would then quantify
into the intensional direct object argument of {\em seek}. As we will
presently see, that is exactly what is needed to derive
{\em
de re} readings.
\subsection{De Re Reading}

We have just seen how theorem \pex{type-raising-theorem} provides a
general mechanism for quantifying into intensional argument
positions. In particular, it allowed the derivation of $\BF{Bill-seeks}'$
from \BF{Bill-seeks}. Now, given the premises
$$\begin{array}[t]{@{}ll@{}}
\BF{Bill-seeks}'\colon & \All{Z}h_\sigma \means Z \linimp f_\sigma
\means \IT{seek}(\IT{Bill}, {\intn}\lam{R}(\extn R)(Z)) \\
\BF{a-unicorn}\colon\ & \All{H, S}(\All{x}
h_{\sigma} \means x \linimp H \means (\extn S)(x))
\linimp H \means \IT{a}(\intn\IT{unicorn}, S)
\end{array}$$
and the variable substitutions
\[\eqalign{Z & \mapsto x \cr
H & \mapsto f_\sigma \cr
S & \mapsto \intn \lam{z}\IT{seek}(\IT{Bill},
{\intn}\lam{R}(\extn R)(z))\cr}
\]
we can apply modus ponens to derive the {\em de re\/} reading of
{\em Bill seeks a unicorn}\/:
\[
f_\sigma \means \IT{a}(\intn\IT{unicorn},
\intn \lam{z}\IT{seek}(\IT{Bill}, {\intn}\lam{R}(\extn R)(z)))
\]

\section{Comparison with Categorial Syntactic Approaches}
\label{sec:categorial}

In recent work, multidimensional and labeled deductive systems
\cite{Moortgat:Labelled,Morrill:type-logical} have been proposed as
refinements of the Lambek systems that are able to represent
synchronized derivations involving multiple levels of representation:
for instance, a level of head-dependent representations and a level of
syntactic functor-argument representations. However, these systems do
not seem yet able to represent the connection between a flat syntactic
representation in terms of grammatical functions, such as the
f-structure of LFG, and a function-argument semantic
representation. The problem in those systems is that they cannot
express at the type level the link between particular syntactic
structures (f-structures in our case) and particular contributions to
meaning. The extraction of meanings from derivations following the
Curry-Howard isomorphism that is standard in categorial systems
demands that the order of syntactic combination coincide with the
order of semantic combination so that functor-argument relations at
the syntactic and semantic level are properly aligned.

Nevertheless, there are strong similarities between the analysis of
quantification that we present and analyses of the same phenomena
discussed by \namecite{Morrill:type-logical} and
\namecite{Carpenter:quant-scope}. Following
\namecite{Moortgat:discontinuous}, they add to an appropriate version
of the Lambek calculus \cite{Lambek:SentStruct} the {\em scope\/}
connective $\Uparrow$, subject to the following proof rules:
\[
\begin{array}{c}
\oneover{\Gamma, v:A, \Gamma'
\Rightarrow u:B\qquad \Delta, t(\lam{v}u):B, \Delta' \Rightarrow C}
{\Delta, \Gamma, t:A\Uparrow B, \Gamma', \Delta' \Rightarrow
C}\quad\mbox{[QL]} \\[1em]
\oneover{\Gamma \Rightarrow u:A}
{\Gamma \Rightarrow \lam{v}v(u):A\Uparrow B}\quad\mbox{[QR]}
\end{array}
\]
\noindent In terms of the scope connective, a quantified NP is given
the category $\mbox{N}\Uparrow\mbox{S}$, which semantically
corresponds to the type \mbox{$(e\rightarrow t)\rightarrow t$} and
agrees with the propositional structure of our linear formulas for
quantified NPs. A phrase of category $N\Uparrow S$ is an
infix functor that binds a variable of type $e$, the type of
individual NPs N, within a scope of type $t$, the type of
sentences S.  An intensional verb like `seek' has, then, category
$(\mbox{N}\setminus(\mbox{S})/(\mbox{N}\Uparrow\mbox{S})$, with
corresponding type $((e\rightarrow t)\rightarrow t)\rightarrow
e\rightarrow t$. \footnote{These category and type assignments are an
oversimplification since intensional verbs like {\em seek}\/ require a
direct object of type $s\rightarrow ((e\rightarrow t)\rightarrow t)$,
but for the present discussion the simpler category and type are
sufficient. \namecite{Morrill:type-logical} provides a full
treatment.} Thus the intensional verb will take as direct object a
quantified NP, as required.

A problem arises, however, with sentences such as
\enumsentence{Bill seeks a conversation with every unicorn.\label{conv}}
This sentence has five possible interpretations:
\eenumsentence{
\item $\IT{seek}(\IT{Bill}, {\intn}\lam{P}
\IT{every}(\intn\IT{unicorn},
\intn\lam{u}\IT{a}(\intn\lam{z}\IT{conv-with}(z,u),P)))$\label{conv-2}
\item $\IT{seek}(\IT{Bill}, {\intn}\lam{P}\IT{a}(\intn\lam{z}
\IT{every}(\intn\IT{unicorn},
\intn\lam{u}\IT{conv-with}(z,u)),P))$\label{conv-1}
\item $\IT{every}(\intn\IT{unicorn},
\intn\lam{u}\IT{seek}(\IT{Bill}, {\intn}\lam{P}
\IT{a}(\intn\lam{z}\IT{conv-with}(z,u),P)))$\label{conv-3}
\item $\IT{every}(\intn\IT{unicorn},
\intn\lam{u}\IT{a}(\intn\lam{z}\IT{conv-with}(z,u),
\intn\lam{z}\IT{seek}(\IT{Bill}, {\intn}\lam{P}(\extn P)(z))))\label{conv-4}$
\item $\IT{a}(\intn\lam{z}
\IT{every}(\intn\IT{unicorn},\intn\lam{u}\IT{conv-with}(z,u)),
\intn\lam{z}\IT{seek}(\IT{Bill}, {\intn}\lam{P}(\extn P)(z)))\label{conv-5}$
}

\noindent Both our approach and the categorial analysis using the scope
connective have no problem in deriving interpretations (\ref{conv-1}),
(\ref{conv-3}), (\ref{conv-4}) and (\ref{conv-5}). In those cases, the
scope of `every unicorn' interpreted as an appropriate term of type
$e\rightarrow t$. However, the situation is different for
interpretation (\ref{conv-2}), in which both the conversations and the
unicorn are {\em de dicto}, but the conversations sought may be
different for different unicorns sought. As we will show below, this
interpretation can be easily derived within our framework. However, a
similar derivation does not appear possible in terms of the categorial
scoping connective.

The difficulty for the categorial account is that the category
$\mbox{N}\Uparrow\mbox{S}$ represents a phrase that plays the role of
a category N phrase where it appears, but takes an S (dependent on the
N) as its scope. In the derivation of (\ref{conv-2}), however, the
scope of `every unicorn' is `a conversation with', which is not of
category S.  Semantically, `a conversation with' is represented
by:
\enumsentence{$
\lam{P}\intn\lam{u}\IT{a}(\intn\lam{z}\IT{conv-with}(z,u),P):
(s\rightarrow e\rightarrow
t)\rightarrow (s\rightarrow e \rightarrow t)$\label{a-conv-type}}
The {\em undirected} Lambek calculus \cite{VanBenthem:LgInAction}
allows us to compose (\ref{a-conv-type}) with the interpretation of
`every unicorn':
\enumsentence{$\lam{Q}
\IT{every}(\intn\IT{unicorn}, Q):(s\rightarrow e\rightarrow t)
\rightarrow t$\label{e-u-typ}}
to yield:
\enumsentence{$\lam{P}
\IT{every}(\intn\IT{unicorn},
\intn\lam{u}\IT{a}(\intn\lam{z}\IT{conv-with}(z,u),P)):(s\rightarrow
e\rightarrow
t)\rightarrow t$\label{e-u-a-c-typ}}
As we will see below, our linear logic formulation also allows that
derivation step.

In contrast, as \namecite{Moortgat:discontinuous} points out, the
categorial rule [QR] is not powerful enough to raise $N\Uparrow S$ to
take as scope any functor whose result is a S. In particular, the
sequent
\enumsentence{$\mbox{N}\Uparrow \mbox{S}\Rightarrow
\mbox{N}\Uparrow(\mbox{N}\Uparrow\mbox{S})\qquad$\label{invalid-seq}}
is not derivable, whereas the corresponding ``semantic'' sequent
(up to permutation)
\enumsentence{$q:(e\rightarrow t)\rightarrow t\Rightarrow$\\
\hspace*{2em} $\lam{R}\lam{P}q(\lam{x}R(P)(x)):((e\rightarrow
t)\rightarrow (e \rightarrow t))\rightarrow (e\rightarrow
t)\rightarrow t$\label{sem-seq}}
is derivable in the undirected Lambek calculus. Sequent
(\ref{sem-seq}) will in particular raise (\ref{e-u-typ}) to a function
that, applied to (\ref{a-conv-type}), produces (\ref{e-u-a-c-typ}), as
required.

Furthermore, the solution proposed by \namecite{Morrill:type-logical}
to make the scope calculus complete is to restrict the intended
interpretation of $\Uparrow$ so that (\ref{invalid-seq}) is not
valid. Thus, {\em contra\/} \namecite{Carpenter:quant-scope}, Morrill's
logically more satisfying account of $\Uparrow$ is not a step towards
making reading (\ref{conv-2}) available.

We now give the derivation of the interpretation (\ref{conv-2}) in our
framework. The f-structure for (\ref{conv}) is:
\enumsentence{\evnup{
\fd{$f$:\fdand{\feat{\pred}{\sc `seek'}
           \feat{\subj}{$g$:\fdand{\feat{\pred}{\sc `Bill'}}}
           \feat{\obj}{$h$:\fdand{\feat{\spec}{\sc `a'}
                                  \feat{\pred}{\sc `conversation'}
				  \feat{\obl\downlett{WITH}}{$i$:
		\fdand{\feat{\spec}{\sc `every'}
		       \feat{\pred}{\sc `unicorn'}}}}}}}}\label{conv-fs}}
The two formulas \BF{Bill-seeks} and \BF{every-unicorn} can be derived
as described before:
$$\begin{array}[t]{ll}
\BF{Bill-seeks}\colon & \All{Y}(\All{s,p}(\All{X}
   h_\sigma \means X \linimp s \means (\extn p)(X))
   \linimp s \means Y(p))\\
   & \qquad \linimp f_\sigma \means \IT{seek}(\IT{Bill}, {\intn}Y)\\[2ex]
\BF{every-unicorn}\colon\ & \All{G, S}
(\All{x}i_{\sigma} \means x \linimp G \means (\extn S)(x))\\
   & \qquad \linimp G \means \IT{every}(\intn\IT{unicorn}, S)
\end{array}$$
The remaining lexical premises for (\ref{conv-fs}) are:
$$\begin{array}[t]{lr@{}l}
\BF{a}\colon &
\All{H, R, T} & (
(\All{x}(h_{\sigma} \var) \means x \linimp (h_{\sigma} \restr) \means
(\extn R)(x))\\
&& \llap{$\otimes$} \/(\All{x} h_{\sigma} \means x \linimp H \means
(\extn T)(x)))\\
&& \qquad \linimp H \means \IT{a}(R, T)\\[2ex]
\BF{conv-with}\colon &
\All{Z, X} &
(h_{\sigma} \var) \means Z \otimes i_\sigma \means X\\
&&  \qquad \linimp  (h_{\sigma} \restr) \means \IT{conv-with}(Z, X)
\end{array}$$
 From these premises we immediately derive
\[
\begin{array}[t]{l}
\All{X, H, T} i_{\sigma} \means X \otimes (\All{x} h_{\sigma} \means x
\linimp H \means (\extn T)(x))\\
\qquad \linimp H \means \IT{a}(\intn\lam{z}\IT{conv-with}(z,X), T)
\end{array}\]
which can be rewritten as:
\enumsentence{$\begin{array}[t]{ll}
\All{H, T}
\lefteqn{(\All{x} h_{\sigma} \means x \linimp H \means (\extn T)(x))
\linimp} \\
& \All{X}(i_{\sigma} \means X \linimp H \means
\IT{a}(\intn\lam{z}\IT{conv-with}(z,X), T))
\end{array}$\label{a-conv-with}}
If we apply the substitutions
\[X  \mapsto x,G \mapsto H,
S \mapsto \intn\lam{u}\IT{a}(\intn\lam{z}\IT{conv-with}(u,v), T),
\]
formula \pex{a-conv-with} can be combined with
\BF{every-unicorn} to yield the required quantifier-type formula:
\enumsentence{$\begin{array}[t]{ll}
\lefteqn{\All{H, T}(\All{x} h_{\sigma} \means x \linimp H \means
(\extn T)(x))
\linimp} \\
& H \means \IT{every}(\intn\IT{unicorn},
\intn\lam{u}\IT{a}(\intn\lam{z}\IT{conv-with}(z,u), T))
\end{array}$\label{a-conv-w-e-un}}
Using substitutions
\[\begin{array}{rcl}
H & \mapsto & s \\
T & \mapsto & p \\
Y & \mapsto &\lam{R}
\IT{every}(\intn\IT{unicorn},
\intn\lam{u}\IT{a}(\intn\lam{z}\IT{conv-with}(z,u),R))
\end{array}
\]
and modus ponens, we then combine \pex{a-conv-w-e-un} with
\BF{Bill-seeks} to obtain the desired final result:
$$
f_\sigma \means
\IT{seek}(\IT{Bill}, {\intn}\lam{R}
\IT{every}(\intn\IT{unicorn},
\intn\lam{u}\IT{a}(\intn\lam{z}\IT{conv-with}(z,u),R)))
$$

\noindent Thus, we see that our more flexible connection between syntax
and semantics permits the full range of type flexibility provided
categorial {\em semantics\/} without losing the rigorous connection to
syntax. In contrast, current categorial accounts of the
syntax-semantics interface do not appear to offer the needed
flexibility when syntactic and semantic composition are more
indirectly connected, as in the present case.

Recently, \namecite{Oehrle:string} independently proposed a
multidimensional categorial system with types indexed so as to keep
track of the syntax-semantic connections that we represent with
$\means$. Using proof net techniques due to
\namecite{Moortgat:Labelled} and \namecite{Roorda:resource},
he maps categorial formulas to first-order clauses similar to our
meaning constructors, except that the formulas arising from
determiners lack the embedded implication. Oehrle's system models
quantifier scope ambiguities in a way similar to ours, but it is not
clear that it can account correctly for the interactions with
anaphora, given the lack of implication embedding in the clausal
representation used.

\section{Conclusion}

Our approach exploits the f-structure of LFG for syntactic information
needed to guide semantic composition, and also exploits the
resource-sensitive properties of linear logic to express the semantic
composition requirements of natural language.  The use of linear logic
as the glue language in a deductive semantic framework allows a
natural treatment of quantification which automatically gives the
right results for quantified NPs, their scopes and bound
anaphora, and allows for a clean and natural treatment of extensional
verbs and their arguments.

Indeed, the same basic facts are also accounted for in other recent
treatments of compositionality, in particular categorial analyses with
discontinuous constituency connectives \cite{Moortgat:discontinuous}.
These results suggest the advantages of a generalized form of
compositionality in which the meaning constructors of phrases are
represented by logical formulas rather than by functional abstractions
as in traditional compositionality. The fixed application order and
fixed type requirements of lambda terms are just too restrictive when
it comes to encoding the freer order of information presentation in
natural language.

In this observation, our treatment is closely related to systems of
syntactic and semantic type assignment based on the Lambek calculus
and its variants.  However, we differ from those categorial approaches
in providing an explicit link between functional structures and
semantic derivations that does not depend on linear order and
constituency in syntax to keep track of predicate-argument relations.
Thus we avoid the need to force syntax and semantics into an
uncomfortably tight categorial embrace.

\section*{Acknowledgments}

Portions of this work were originally presented at the Second CSLI
Workshop on Logic, Language, and Computation, Stanford University, and
published as \namecite{DLPS:QuantLFG}; and at the Conference on
Information-Oriented Approaches to Logic, Language and Computation,
held at Saint Mary's College, Moraga, California, and published as
\namecite{DLPS:Intensional}.  We are grateful to the audiences at
these venues for helpful comments.  We would particularly like to
thank Johan van Benthem, Bob Carpenter, Jan van Eijck, Kris Halvorsen,
Angie Hinrichs, David Israel, Ron Kaplan, Chris Manning, John Maxwell,
Michael Moortgat, John Nerbonne, Stanley Peters, Henriette de Swart
and an anonymous reviewer for the Second CSLI Workshop on Logic,
Language, and Computation for comments and discussion.  They are not
responsible for any remaining errors, and we doubt that they will
endorse all our analyses and conclusions, but we are sure that the end
result is much improved for their help.

\newpage
\appendix
\section{Syntax of the Meaning and Glue Languages}
\label{sec:syn-app}

The meaning language is based on Montague's
intensional higher-order logic, with the following syntax:
$$\begin{array}{lrcll}
\mbox{(M-terms)} & M & {::=}  & c & \mbox{(Constants)}\\
& & \alt & x & \mbox{(Lambda-variables)} \\
& & \alt & \lambda x\, M & \mbox{(Abstraction)}\\
& & \alt & M\, M & \mbox{(Application)}\\
& & \alt & X & \mbox{(Glue-language variables)} \\
& & \alt & \hat{\ } & \mbox{(``cap'' operator)} \\
& & \alt & \check{\ } & \mbox{(``cup'' operator)} \\
\end{array}$$

Terms are typed in the usual way; logical connectives such as
{\em every\/} and {\em a\/} are represented by constants of
appropriate type. The ``cap'' operator is polymorphic, and of type
$\alpha \rightarrow (s \rightarrow \alpha)$; similarly the
``cup'' operator is of type $(s \rightarrow \alpha) \rightarrow
\alpha)$.

For readability, we will often ``uncurry'' $M N_1 \cdots N_m$ as
$M(N_1, \ldots, N_m)$. Note that we allow variables in the glue
language to range over meaning terms.

The glue language refers to three kinds of terms: meaning terms,
f-structures, and semantic or $\sigma$-structures. f- and
$\sigma$-structures are feature structures in correspondence (through
projections) with constituent structure. Conceptually, feature
structures are just functions which, when applied to attributes (a set
of constants), return constants or other feature structures.  In the
following we let $A$ range over some pre-specified set of attributes.
$$\begin{array}{lrcll}
\mbox{(F-terms)} & F & {::=}  & \uparrow & \mbox{(Indexical
reference)}\\
& & \alt & f\; \alt\; g \;\alt\; h\; \alt \cdots & \mbox{(F-structure
constants)} \\
& & \alt & (F A) & \mbox{(Attribute selection)}\\
\quad\\
\mbox{($\sigma$-terms)} & S & {::=}
& F_{\sigma} & \mbox{(Semantic projection)}\\
& & \alt & (S A) & \mbox{(Attribute selection)}\\
& & \alt & H & \mbox{(Glue-language variable)}
\end{array}$$

Glue-language formulas are built up using linear connectives from
atomic formulas of the form $S \means_{\tau} M$, whose intended
interpretation is that the meaning associated with $\sigma$-structure
$S$ is denoted by term $M$ of type $\tau$. We omit the type subscript
$\tau$ when it can be determined from context.
$$\begin{array}{lrcll}
\mbox{(Glue formulas)} & G & {::=}  &S \means_{\tau} M &
\mbox{(Basic assertion)}\\
& & \alt & G \otimes G & \mbox{(Linear conjunction)}\\
& & \alt & G \linimp G & \mbox{(Linear implication)}\\
& & \alt & \Pi \lambda X.\,G  & \mbox{(Quantification over M-terms)}\\
& & \alt & \Pi \lambda H.\,G  & \mbox{(Quantification over $\sigma$-terms)}\\
\end{array}$$
We usually write $\Pi \lambda X.\,G$ as $\forall X.\,G$, and
similarly for $\Pi \lambda H.\,G$.

\newpage
\section{Proof rules for intensional higher-order linear logic}
\label{sec:rules-app}
\def\la{\linimp}
\begin{tabular}{rccl}
{\bf Identity} &   $\begin{array}{c}
	            \hline
		    F \vdash F
		    \end{array}$
         &  $\begin{array}{c}
	    \Gamma_1 \vdash F \hspace{1cm}
            \Gamma_2, F \vdash D\\
            \hline
            \Gamma_1, \Gamma_2 \vdash D
            \end{array}$ & {\bf Cut} \\[.5cm]
{\bf Exch. Left} &  $\begin{array}{c}
	    \Gamma_1, F, G, \Gamma_2 \vdash D \\
            \hline
            \Gamma_1, G, F, \Gamma_2 \vdash D
            \end{array}$
         &  \\[.5cm]
{\bf $\lambda$ Left} &   $\begin{array}{c}
	    \Gamma, F' \vdash D \hspace{1cm}
             F \rightarrow_{\lambda} F'\\
            \hline
	    \Gamma, F \vdash D
		    \end{array}$
         &  $\begin{array}{c}
	    \Gamma \vdash D \hspace{1cm}
             D \rightarrow_{\lambda} D'\\
            \hline
	    \Gamma \vdash D'
            \end{array}$ & {\bf $\lambda$ Right} \\[.5cm]
{\bf $\otimes$ Left} &  $\begin{array}{c}
	    \Gamma, F, G \vdash D \\
            \hline
            \Gamma, (F \otimes G) \vdash D
            \end{array}$
         &  $\begin{array}{c}
	    \Gamma_1 \vdash F \hspace{1cm}
            \Gamma_2 \vdash G\\
            \hline
            \Gamma_1, \Gamma_2 \vdash (F \otimes G)
            \end{array}$ & {\bf $\otimes$ Right} \\[.5cm]
{\bf $\la$ Left} &  $\begin{array}{c}
	    \Gamma_1 \vdash F \hspace{1cm}
            \Gamma_2, G \vdash D\\
            \hline
            \Gamma_1, \Gamma_2, (F \la G) \vdash D
            \end{array}$
         &  $\begin{array}{c}
	    \Gamma, F \vdash G \\
            \hline
            \Gamma \vdash (F \la G)
            \end{array}$ & {\bf $\la$ Right} \\[.5cm]
{\bf $\Pi$ Left} & $\begin{array}{c}
	     \Gamma,P t\vdash D \\
             \hline
	     \Gamma,\Pi P \vdash D \\
             \end{array}$
           & $\begin{array}{c}
	     \Gamma  \vdash P y \\
             \hline
	     \Gamma  \vdash \Pi P \\
             \end{array}$ & {\bf $\Pi$ Right} \\[.5cm]
\end{tabular}

The {\bf $\Pi$ Right} rule only applies if $y$ is not free in
$\Gamma,\Sigma$, and any nonlogical theory axioms. We write $M
\rightarrow_{\lambda} N$ to indicate that $N$ can be obtained
from $M$ by one or more applications of $\alpha-$ or $\beta-$
reduction, or by the application of the rule:
$$\extn(\intn(Q)) \rightarrow Q$$
to a sub-term of $M$.

\newpage

\end{document}